\documentclass[10pt,journal,compsoc]{IEEEtran}
\ifCLASSOPTIONcompsoc
  \usepackage[nocompress]{cite}
\else
  \usepackage{cite}
\fi
\usepackage{booktabs}
\usepackage{epsfig}
\usepackage{graphicx}
\usepackage{amsmath}
\usepackage{amssymb}

\usepackage{url}
\usepackage{tabularx}
\usepackage{multirow}
\usepackage{subfigure}
\usepackage{amsmath,dsfont}
\usepackage{array}
\usepackage{enumitem}

\usepackage{pifont}
\usepackage{threeparttable}

\usepackage{verbatim}

\usepackage{colortbl}
\definecolor{mygray}{gray}{.75}

\usepackage{xcolor}

\usepackage[normalem]{ulem}

\newcommand{\argmin}{\operatornamewithlimits{argmin}}

\newcommand{\eg}{\emph{e.g.,}~}
\newcommand{\etal}{\emph{et~al.}~}
\newcommand{\ie}{\emph{i.e.,}~}

\ifCLASSINFOpdf
\else
\fi
\begin{document}
%
\title{Dual Encoding for Video Retrieval by Text}
%
%
%
%

\author{Jianfeng~Dong,
        Xirong~Li,~\IEEEmembership{Member,~IEEE,}
        Chaoxi~Xu,
        Xun~Yang,
        Gang~Yang,\\
        Xun~Wang,~\IEEEmembership{Member,~IEEE,}
        Meng~Wang,~\IEEEmembership{Fellow,~IEEE,}
\IEEEcompsocitemizethanks{
\IEEEcompsocthanksitem J. Dong and X. Wang are with the College of Computer and Information Engineering, Zhejiang Gongshang University, Hangzhou 310035, China. \protect\\
E-mail: dongjf24@gmail.com
\IEEEcompsocthanksitem X. Li, C. Xu and G. Yang are with the Key Lab of Data Engineering and Knowledge Engineering, Renmin University of China, and the AI and Media Computing Lab, School of Information, Renmin University of China, Beijing 100872, China.\protect\\
E-mail: xirong@ruc.edu.cn
\IEEEcompsocthanksitem X. Yang is with the School of Computing, National University of Singapore, Singapore 37580, Singapore.\protect\\
E-mail: xunyang@nus.edu.sg
\IEEEcompsocthanksitem M. Wang is with the School of Computer Science and Information Engineering, Hefei University of Technology, Hefei 230009, China \protect\\
E-mail: wangmeng@hfut.edu.cn
}
\thanks{Manuscript received Sep 2, 2020; revised Dec 30, 2020 and Feb 9, 2021; accepted Feb 10, 2021. (Corresponding authors: Xirong Li and Xun Wang)}}

%
%

\markboth{IEEE Transactions on Pattern Analysis and Machine Intelligence,~Vol.~, No.~, February~2021}%
{Dong \MakeLowercase{\textit{et al.}}: Dual Encoding for Video Retrieval by Text}
%



\IEEEtitleabstractindextext{%
\begin{abstract}
This paper attacks the challenging problem of video retrieval by text. In such a retrieval paradigm, an end user searches for unlabeled videos by ad-hoc queries described exclusively in the form of a natural-language sentence, with no visual example provided. Given videos as sequences of frames and queries as sequences of words, an effective sequence-to-sequence cross-modal matching is crucial. To that end, the two modalities need to be first encoded into real-valued vectors and then projected into a common space. In this paper we achieve this by proposing a \textit{dual} deep \textit{encoding} network that encodes videos and queries into powerful dense representations of their own. Our novelty is two-fold. First, different from prior art that resorts to a specific single-level encoder, the proposed network performs multi-level encoding that represents the rich content of both modalities in a coarse-to-fine fashion. Second, different from a conventional common space learning algorithm which is either concept based or latent space based, we introduce hybrid space learning which combines the high performance of the latent space and the good interpretability of the concept space. Dual encoding is conceptually simple, practically effective and end-to-end trained with hybrid space learning. Extensive experiments on four challenging video datasets show the viability of the new method. Code is available at \url{https://github.com/danieljf24/hybrid_space}.
\end{abstract}

\begin{IEEEkeywords}
Video retrieval, cross-modal representation learning, dual encoding, hybrid space learning
\end{IEEEkeywords}}

\maketitle

\IEEEdisplaynontitleabstractindextext

%
\IEEEpeerreviewmaketitle

\IEEEraisesectionheading{\section{Introduction}\label{sec:introduction}}

\IEEEPARstart{T}{his} paper targets at the task of video retrieval by text, where a query is described exclusively in the form of a natural-language sentence, with no visual example attached. The task is scientifically interesting and challenging as it requires establishing proper associations between visual and linguistic information presented in the temporal order. 

Retrieving \emph{unlabeled} videos by text attracts initial attention in the form of zero-example multimedia event detection, where the goal is to retrieve video shots showing specific events such as \emph{parking a vehicle}, \emph{dog show} and \emph{birthday party}, but with no training videos provided \cite{cikm13-zsvr,mm14-cmu-jiang,icmr14-amir,cvpr14-wu-zsed,aaai15-zsed,icmr16-zsed,pami2017-videostory}. All these methods are concept based, representing the video content by automatically detected concepts, which are used to match with textual descriptions of a target event. An attractive property of the concept-based representation is its good interpretability~\cite{pami2017-videostory}, as each of its dimensions has explicit meanings. The concept-based tradition continues in the era of deep learning. For the NIST TRECVID AVS challenge~\cite{AwadTRECVID16,icmr17-awad}, a leading benchmark evaluation for video retrieval by text, we observe that the top performers in the previous years (2016--2018) are mostly concept based \cite{tv16-nii,tv16-certh,tv17-waseda,tv17-vireo,tv18-waseda}. However, the concept-based paradigm faces intrinsic difficulties including how to specify a set of proper concepts, how to train good classifiers for these concepts, and more crucially how to select relevant and detectable concepts for both video and query representation \cite{icmr16-zsed}. These difficulties remain largely unresolved to this day.

\begin{figure}[tb!]
\centering\includegraphics[width=\columnwidth]{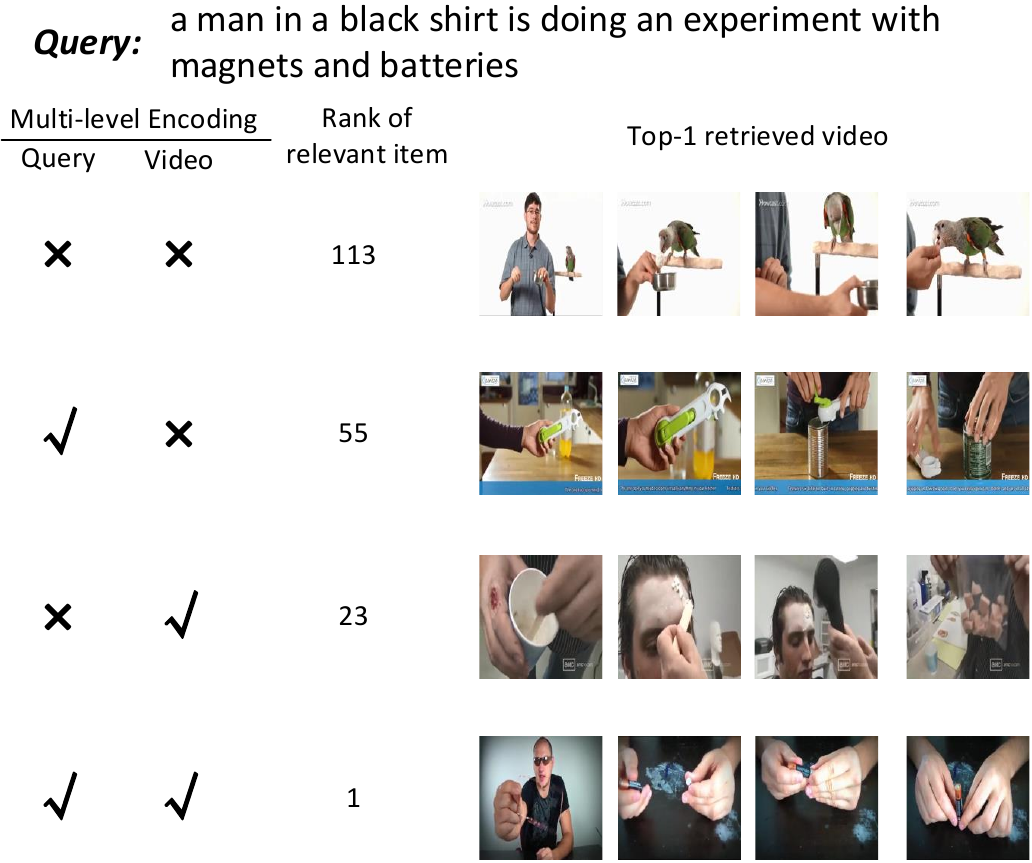}
\caption{\textbf{Showcase of video retrieval by text with and without the proposed encoding}. The \ding{54} symbol indicates encoding by mean pooling. Numbers in the third column are the rank of the relevant video returned by retrieval models subject to specific query / video encoding strategies. The retrieval model with dual encoding successfully answers this complex query. }\label{fig:showcase}
\end{figure}

Not surprisingly, efforts towards concept-free representation have been made. In the context of cross-modal retrieval between image and text~\cite{pami2014-pereira,wu2013cross}, Canonical Correlation Analysis (CCA) has been frequently used to linearly project both visual and textual features into a common space. From the viewpoint of neural networks, CCA essentially applies a fully connected (FC) layers (with no bias term) on the visual side and another FC layer on the textual side. With the quick development of deep neural networks in both computer vision and natural language processing, the simple FC layer has now been replaced by more advanced embedding networks~\cite{mithun2018learning,zhang2018cross,liu2019use,miech2019howto100m,sigir2020tree,chen2020fine}.  
Still, we see a common pattern among the varied solutions. That is, first encoding videos and textual queries, and then mapping them into a common space where the video-text similarities can be computed directly. Therefore, what matters for this paradigm are forms of video encoding, text encoding and common space learning.

For video encoding, a popular solution is to first extract frame features from videos by pre-trained CNN models, and then aggregate them into a video-level feature by mean pooling \cite{eccv2016ws-otani,mithun2018learning}, max pooling\cite{wray2019fine,miech2019howto100m}, NetVLAD \cite{liu2019use}, RNN \cite{dongdl,torabi2016learning} or self-attention mechanisms \cite{sigir2020tree,chen2020fine}. 
As for text encoding, while bag-of-words remains common \cite{pami2017-videostory,kratochvil2020som}, deep networks are in an increasing use. Given a specific sentence, a typical approach is to first quantize each of its word by word embedding and then aggregate word-level features into a sentence-level feature by max pooling \cite{miech2019howto100m}, Fisher Vector \cite{miech2018learning,shao2018find}, NetVLAD \cite{wray2019fine,liu2019use}, RNN \cite{yu2018joint,mithun2018learning} or Graph Convolutional Network (GCN) \cite{chen2020fine}.  
Instead of using one specific encoding strategy, Dong \etal \cite{dong2018predicting} and Li \etal \cite{li2019w2vv++} utilize multiple text encoders including bag-of-words, word2vec and GRU. However, they simply use mean pooling for video encoding. 
In contrast to the existing works, we propose \emph{dual multi-level encoding} for both videos and text in advance to common space learning. As exemplified in Fig. \ref{fig:showcase}, the new encoding strategy is crucial for describing complex queries and video content.

Our hypothesis is that a given video/query has to be first encoded into a powerful representation of its own. We consider such a decomposition crucial as it allows us to design an encoding network that jointly exploits multiple encoding strategies including mean pooling, recurrent neural networks and convolutional networks.
In our design, the output of a specific encoding block is not only used as input of a follow-up encoding block, but also re-used via skip connections to contribute to the final output. 
It generates new, higher-level features progressively. These features, generated at distinct levels, are powerful and complementary to each other, allowing us to obtain effective video (and text) representations by very simple concatenation.

Philosophically, our dual encoding model is linked to Allan Paivio's dual-coding theory of cognition~\cite{paivio1991dual}. Supported by evidence from psychological research, the dual-coding theory postulates that verbal and visual information are processed along distinct channels with separate representations in the human mind. Later, these representations are used for retrieving information previously stored in the mind. In a similar spirit, the dual encoding model stores video and textual information learned from training data in separate representations and recall them in the inference stage.

For common space learning, the state-of-the-art relies on constructing a latent space~\cite{sigir2020tree,chen2020fine,liu2019use,miech2019howto100m}, as such a space can be optimized in an end-to-end manner, and thus permits superior performance against the concept-based alternative. 
This, however, comes at the cost of losing interpretability. Different from the concept space, each dimension of the latent space is not directly interpretable. Hence, what a model has truly learned is often agnostic. 
In order to combine the merits of the latent space and the concept space, we propose to train the dual encoding network with \textit{hybrid space} learning. In particular, a latent space and a concept space are simultaneously learned, for better performance and better interpretability.

In sum, this paper makes the following contributions. 
\begin{itemize}
	\item We propose a novel \emph{dual} network that encodes an input, let it be a query sentence or a video, in a similar manner. By jointly exploiting multi-level encodings, the network explicitly and progressively learns to represent global, local and temporal patterns in videos and sentences. Moreover, dual encoding is orthogonal to common space learning, allowing us to flexibly embrace state-of-the-art common space learning algorithms. 
	\item We propose a new hybrid space learning to learn a hybrid common space for  video-text similarity prediction, which inherits the high performance of the latent space and the good interpretability of the concept space. 
	\item We conduct extensive experiments on four challenging video datasets, \ie MSR-VTT~\cite{xu2016msr}, TRECVID AVS 2016-2018~\cite{AwadTRECVID16,AwadTRECVID17,AwadTRECVID18}, VATEX~\cite{wang2019vatex} and MPII-MD~\cite{rohrbach2015dataset}. Dual encoding, trained by hybrid space learning, is a new state-of-the-art for video retrieval by text. 
\end{itemize}

The architecture of our model is based on existing components including mean feature pooling, GRU and 1D-CNN. Our novelty is the integration of these components into a dual multi-level encoding architecture for video / text representation learning. We consider the novelty of the system architecture to be more than the sum of its parts. The Dual Encoding network effectively combines these vanilla components into a powerful solution that is competitive with recent Transformer based methods ~\cite{zhu2020actbert,gabeur2020multi}.
Also notice that while we use two pre-trained CNNs to extract frame-level features, the two features are concatenated to form a single vector before feeding it into our network. The multi-level features combined in Dual Encoding are generated by the network itself. Hence, our method essentially uses a single visual feature as input, and thus differs fundamentally from previous works that rely on the ensemble of diverse models / features.

A preliminary version of this work was published at CVPR 2019 \cite{cvpr2019-dual-dong}. 
The journal extension improves over the conference paper mainly in two aspects. Technically, we introduce hybrid space learning. Compared to the latent space learning used in \cite{cvpr2019-dual-dong}, the new learning algorithm leads to better video retrieval performance. With the learned concept space, the model's interpretability is also improved. Experimentally, our evaluation has been substantially expanded in terms of datasets and competitive baselines.

\section{Related Work} \label{sec:rel-work}

\subsection{Concept based Methods}
Since 2016 the TRECVID starts a new challenge for video retrieval by text, known as Ad-hoc Video Search (AVS) \cite{AwadTRECVID16}.
The majority of the top ranked solutions for this challenge depend on visual concept classifiers to describe video content and linguistic rules to detect concepts in textual queries \cite{tv16-nii,tv16-certh,icmr2017-certh-avs,tv17-waseda,tv17-vireo}.
Then the similarity between a textual query and a specific video is typically computed by concept matching. 
For instance, \cite{tv16-certh,icmr2017-certh-avs} utilize multiple pre-trained Convolutional Neural Network (CNN) models to detect main objects and scenes in video frames. As for query representation, the authors design relatively complex linguistic rules to extract relevant concepts from a given query. Ueki \etal \cite{tv17-waseda} come with a much larger concept bank consisting of more than 50k concepts. In addition to pre-trained CNN models, they train SVM classifiers to automatically annotate the video content.
In \cite{tv17-uva}, Snoek \etal utilize a model called VideoStory \cite{pami2017-videostory} to represent videos and then embed them into a concept space by a linear transformation, while they still represent textual query by selecting concepts based on part-of-speech tagging heuristically.  Consequently, the video-text similarity is implemented as the cosine similarity in terms of their concept vectors.

We argue that such a concept-based paradigm has a fundamental disadvantage. That is, it is very difficult to describe the rich sequential information within both video and query using a few selected concepts. Despite the disadvantage, such a paradigm also has its merit where representing video and textual query by concepts make it somewhat interpretable.
In this work, we also integrate such interpretation merit into our proposed model, allowing the model to match videos and text in the concept space.

\subsection{Latent Space based Methods}
Latent space based methods first encode video and textual queries and then map them into a common latent space where the video-text similarity can be measured directly~\cite{aaai2015-xu-video,mithun2018learning,xiong2019graph,li2019w2vv++}. For these methods, what matters are forms of video encoding, text encoding and similarity learning. So we review recent work in these aspects.

For video encoding, a typical approach is to first extract visual features from video frames by pre-trained CNN models, and subsequently aggregate the frame-level features into a video-level feature. The \emph{de facto} choice is mean pooling \cite{eccv2016ws-otani,dong2018predicting,mithun2018learning,shao2018find,liu2019use} or max pooling \cite{miech2018learning,wray2019fine,miech2019howto100m}. 
To explicitly model the temporal information, Torabi \etal \cite{torabi2016learning} use a Long Short-Term Memory (LSTM), where frame-level features are sequentially fed into the LSTM, and the mean pooling of the hidden vectors at each step is used as the video feature.
Besides using a Gated Recurrent Unit (GRU) to model the temporal dependency of video frames, Yang \etal \cite{sigir2020tree} additionally utilize the multi-head self-attention mechanism \cite{vaswani2017attention} to learn the frame-wise correlation thus enhance the video representation.
In addition to the frame-level visual features, we also notice efforts on utilizing features extracted from other channels such as audio \cite{liu2019use,mithun2018learning,miech2018learning} and motion \cite{miech2019howto100m,liu2019use,mithun2018learning,miech2018learning}. However, they still use mean pooling, max pooling or NetVLAD to aggregate different features into video-level features.

For text encoding, while bag-of-words (BoW) remains common \cite{habibian2014videostory,pami2017-videostory,kratochvil2020som}, deep networks are in increasing use. 
The common way by deep learning techniques is to first represent each word of the textual query by word2vec models pre-trained on large-scale text corpora, and then aggregate them by max pooling in \cite{miech2019howto100m}, Fisher Vector in \cite{miech2018learning,shao2018find} or NetVLAD in \cite{wray2019fine,liu2019use}. Despite their good performance, the main drawback of these methods is ignoring the sequential order in text.
Recurrent neural networks (RNN), known to be effective for modeling sequence oder dependency, are also dominated \cite{aaai2015-xu-video,yu2017end,yu2018joint,mithun2018learning,eccv2016ws-otani}. 
Recursive neural networks are investigated in \cite{aaai2015-xu-video} for vectorizing subject-verb-object triplets extracted from a given sentence. Variants of recurrent neural networks are being exploited, see the usage of LSTM, bidirectional LSTM, GRU, bidirectional GRU in \cite{yu2017end}, \cite{yu2018joint}, \cite{mithun2018learning} and \cite{song2019polysemous}, respectively. 
For instance, Mithun \etal \cite{mithun2018learning} utilize the last hidden state of the GRU as the text representation.
The work \cite{dong2018predicting} and its extension \cite{li2019w2vv++} explore a joint use of multiple text encoding strategies including BoW, word2vec and GRU, and found it is beneficial for video retrieval. However, as aforementioned, those works simply employ mean pooling for video encoding. 
Recently, Chen \etal \cite{chen2020fine} utilize graph convolutional network to model the connection between words. But it requires text to be well annotated with semantic role relation annotations, which may be not suitable for a new scenario with different linguistic expression patterns.
By contrast, our proposed text encoding trained in an end-to-end manner, without using such semantic role annotations. Moreover, different from the above works, this paper aims to explicitly and progressively exploit global, local and temporal patterns in both videos and textual queries.

For similarity learning, lots of works \cite{aaai2015-xu-video,eccv2016ws-otani,dong2018predicting,li2019w2vv++,miech2019howto100m,liu2019use} typically project encoded videos and text into a common latent space and triplet ranking loss variants are dominantly used for model training.  After being projected in a latent space, the video-text similarity can be measured by a standard similarity metric, \eg cosine similarity. 
Besides the triplet ranking loss, Zhang \etal \cite{zhang2018cross} additionally employ contrastive loss and reconstruction loss to further constrain the latent space.
Recently, we notice an increasing used of learning multiple common latent spaces \cite{mithun2018learning,miech2018learning,wray2019fine,chen2020fine} instead of only one latent space. For instance, Miech \etal \cite{miech2018learning} utilize four different features, \ie appearance, motion, face and audio features, to represent videos, and learn one latent space for each video feature. The final video-text similarity is the fusion of their similarities in the four latent spaces.  Wray \etal \cite{wray2019fine} decompose text into nouns and non-noun words, and respectively project them into two different latent spaces.
With the similar idea of \cite{wray2019fine}, Chen \etal \cite{chen2020fine} decompose text into events, actions and entities, and three latent spaces are learned for video-text matching.

Different from the existing works where learned common spaces are latent spaces without no explicit interpretability, we propose to learn a hybrid space consisting of a latent space and a concept space, which inherits the merits of high performance of latent space based methods and interpretability of concept based methods.

\begin{figure*}[tb!]
\centering\includegraphics[width=2\columnwidth]{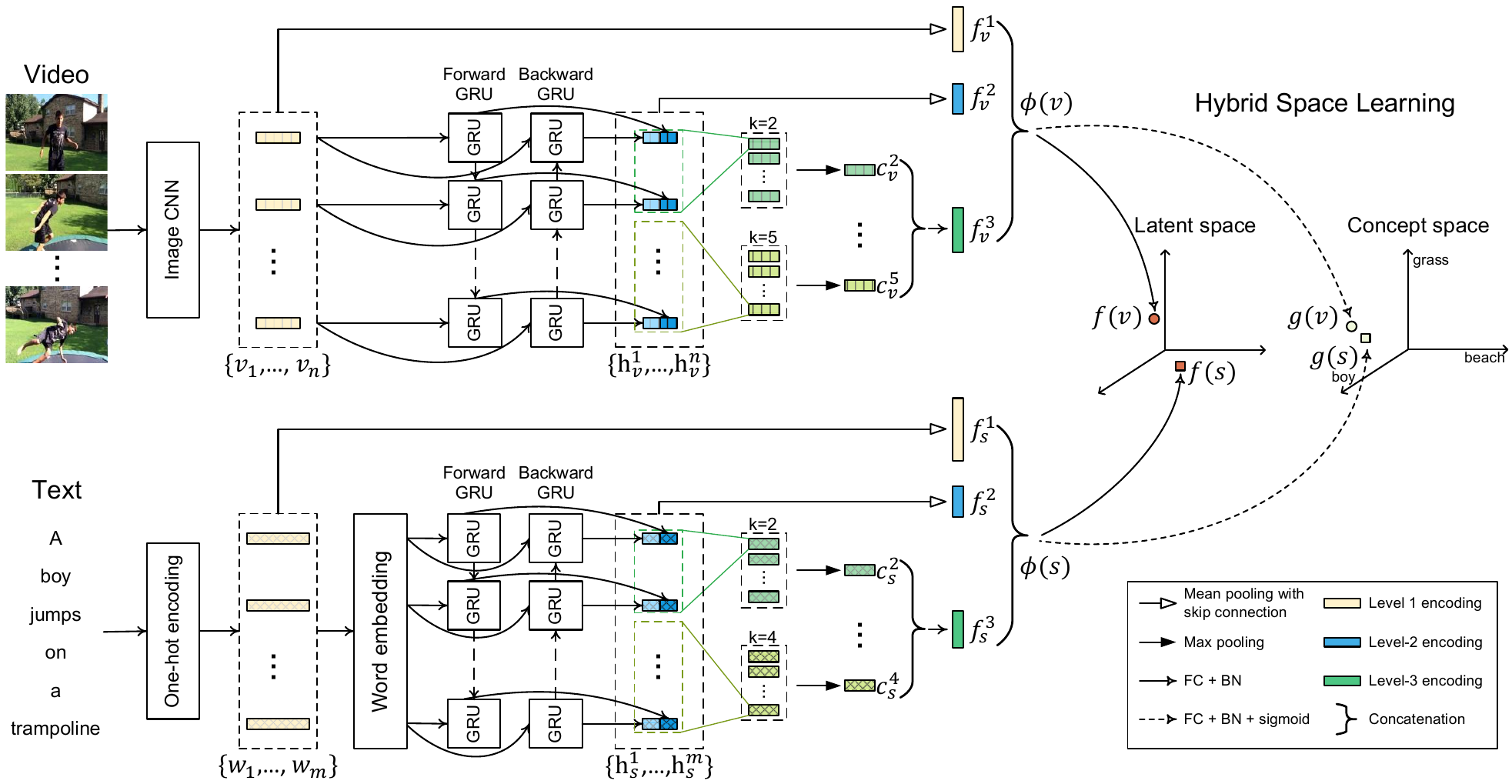}
\vspace{-4mm}
\caption{\textbf{A conceptual diagram of the proposed dual encoding network for video retrieval by text.} Given a video $v$ and a sentence $s$, the network performs in parallel multi-level encodings, \ie mean pooling, biGRU and biGRU-CNN, eventually representing the two input by two combined vectors $\phi(v)$ and $\phi(s)$, respectively. The vectors are later projected into a hybrid common space which consists of a latent space and a concept space. Once the network is trained, encoding at each side is performed independently, meaning we can process large-scale videos offline and answer ad-hoc queries on the fly.}\label{fig:framework}
\end{figure*}

\subsection{Cross-modal Fusion Methods}
In contrast to latent space based methods, cross-modal fusion methods do not construct an explicit latent space.
Instead, they use a cross-modal fusion subnetwork that takes text and videos as input and directly produces similarity scores \cite{yu2018joint,yu2017end,zhu2020actbert}.
For instance, Yu \etal \cite{yu2018joint} fuse the video and text representation into a 3D tensor using a soft attention, and a convolutional network is further employed to directly predict the similarity based on the fused feature. 
Although these methods \cite{yu2018joint,yu2017end} are effective, their retrieval efficiency are somewhat low as video and text are coupled with each other. By contrast, our proposed method maps videos and text into a common space by a separated branch respectively, which makes videos and text decoupled. So, all candidate videos can be mapped in the common space off-line, which is efficient for large-scale video retrieval.

\subsection{Transformer-based Methods}
Transformers~\cite{vaswani2017attention} are a new genre of deep neural networks that stack multiple self-attended layers. By self-supervised learning with masked language modeling on very large-scale corpora, BERT-like Transformer models have shown superior performance over RNNs on multiple NLP tasks~\cite{devlin2018bert}. Unsurprisingly, we see an increasing effort on re-purposing such types of models for video retrieval by text ~\cite{gabeur2020multi,zhu2020actbert,lokoc2020w2vvpp,zhao2020stacked}. In Loko{\'c} \etal \cite{lokoc2020w2vvpp} and  Zhao \etal \cite{zhao2020stacked}, a pre-trained BERT is adopted to encode the textual query. Gabeur \etal \cite{gabeur2020multi} also use BERT for query representation, and go one step further by proposing a multi-modal Transformer (MMT) with four stacked Transformer layers to jointly encode diverse video features for video representation. 
Zhu and Yang propose ActBERT to encode global actions, local regional objects, and text descriptions in a unified framework~\cite{zhu2020actbert}.
We empirically show in Section \ref{sssec:trans} that our proposed method, although utilizing relatively simple encoders such as Bag-of-Words, GRU and 1D-CNN, compares favorably to the Transformer based methods. Moreover, our method can be extended with ease to incorporate the Transformers for better performance.

\section{The \emph{Dual Encoding} Network} \label{sec:method}

Both video and sentence are essentially a sequence of items, let it be frames or words. Such a property motivates us to design a dual encoding network to handle the two distinct modalities. Specifically, given a video $v$ and a sentence $s$, the proposed network encodes them in parallel, in advance to common space learning. As illustrated in Fig. \ref{fig:framework}, multi-level encodings are performed for each modality. The encoding results are then combined, denoted as $\phi(v)$ and $\phi(s)$, to describe the two modalities in a coarse-to-fine fashion. In our design, $\phi(v)$ and $\phi(s)$ are not directly used for cross-modal matching. So the two vectors do not have to reside in the same feature space and can have distinct dimensions, giving them sufficient freedom to become powerful representations of the corresponding modalities. 

In what follows we first depict the network at the video side. We then specify choices that are unique at the text side.

\subsection{Video-side Multi-level Encoding}

For a given video, we extract uniformly a sequence of $n$ frames with a pre-specified interval of 0.5 seconds. Per frame we extract deep features using a pre-trained ImageNet CNN, as commonly used for video content analysis \cite{aaai2015-xu-video,pami2017-videostory,icmr2017-certh-avs}. Consequently, the video is described by a sequence of feature vectors $\{v_1,v_2,\ldots,v_n\}$, where $v_t$ indicates the deep feature vector of the $t$-th frame.  Notice that 3D CNNs \cite{tran2015learning,carreira2017quo} can also be used for feature extraction when treating segments of frames as individual items.

\subsubsection{Level 1. Global Encoding by Mean Pooling}
According to our literature review, mean pooling, which represents a video by simply averaging the features of its frames, is arguably the most popular choice for text-video retrieval. By definition, mean pooling captures visual patterns that repeatedly present in the video content. These patterns tend to be global. We use $f_v^{1}$ to indicate the encoding result at this level, that is:
\begin{equation}
f_v^{1} = \frac{1}{n}\sum_{t=1}^n v_t.
\end{equation}

\subsubsection{Level 2. Temporal-Aware Encoding by biGRU}
Bi-directional recurrent neural network \cite{birnn} is known to be effective for making use of both past and future contextual information of a given sequence. We hypothesize that such a network is also effective for modeling the video temporal information. We adopt a bidirectional GRU (biGRU) \cite{cho2014learning}, which has less parameters than the bidirectional LSTM and thus requires less amounts of training data.
A biGRU consists of two separated GRU layers, \ie a forward GRU and a backward GRU. The forward GRU is used to encode frame features in normal order, while the backward GRU encodes frame features in reverse order. Let $\overrightarrow{h}_t$ and $\overleftarrow{h}_t$ be their corresponding hidden states at a specific time step $t=1,\ldots,n$. The hidden states are generated as
\begin{equation}
\begin{array}{l}
\overrightarrow{h}_t = \overrightarrow{GRU}(v_t, \overrightarrow{h}_{t-1}), \\
\overleftarrow{h}_t = \overleftarrow{GRU}(v_{n+1-t}, \overleftarrow{h}_{t-1}), 
\end{array} 
\end{equation}
where $\overrightarrow{GRU}$ and $\overleftarrow{GRU}$ indicate the forward and backward GRUs, with past information carried by $\overrightarrow{h}_{t-1}$ and $\overrightarrow{h}_{t-1}$, respectively. Concatenating $\overrightarrow{h_{t}}$ and $\overleftarrow{h_{t}}$, we obtain the biGRU output $h_v^t = [\overrightarrow{h}_t, \overleftarrow{h}_t]$. The size of the hidden vectors in the forward and backward GRUs is empirically set to 512. Accordingly, the size of $h_v^t$ is 1,024. Putting all the output together, we obtain a feature map $H_v=\{h_v^1, h_v^2,..., h_v^n\}$, with a size of $1,024 \times n$. The biGRU based encoding, denoted $f_v^{(2)}$, is obtained by applying mean pooling on $H_v$ along the row dimension, that is
\begin{equation}
f_v^{2} = \frac{1}{n}\sum_{t=1}^n h_v^t.
\end{equation}

\subsubsection{Level 3. Local-Enhanced Encoding by\\ biGRU-CNN}
The previous layer treats the output of biGRU at each step equally. To enhance local patterns that help discriminate between videos of subtle difference, we build convolutional networks on top of biGRU. In particular, we adapt 1D CNN originally developed for sentence classification \cite{kim2014convolutional}. 

The input of our CNN is the feature map $H_v$ generated by the previous biGRU module. Let $Conv1d_{k,r}$ be a 1D convolutional block that contains $r=512$ filters of size $k$, with $k \ge 2$. Feeding $H_v$, after zero padding, into $Conv1d_{k,r}$ produces a $n \times r$ feature map. Non-linearity is introduced by applying the ReLU activation function on the feature map. As $n$ varies for videos, we further apply max pooling to compress the feature map to a vector $c_k$ of fixed length $r$. More formally we express the above process as 
\begin{equation}
c_v^k = \mbox{max-pooling}(ReLU(Conv1d_{k,r}(H_v))).
\end{equation}

A filter with $k=2$ allows two adjacent rows in $H_v$ to interact with each other, while a filter of larger $k$ means more adjacent rows are exploited simultaneously. In order to generate a multi-scale representation, we deploy multiple 1D convolutional blocks with $k=2,3,4,5$. Their output is concatenated to form the biGRU-CNN based encoding, \ie
\begin{equation}
f_v^{3} = [c_v^2, c_v^3, c_v^4, c_v^5].
\end{equation}

As $f_v^{1}$, $f_v^{2}$, $f_v^{3}$ are obtained sequentially at different levels by specific encoding strategies, we consider it reasonable to presume that the three encoding results are complementary to each other, with some redundancy. Hence, we obtain multi-level encoding of the input video by concatenating the output from all the three levels, namely
\begin{equation}
\phi(v) = [f_v^{1}, f_v^{2}, f_v^{3}].
\end{equation}
In fact, this concatenation operation, while being simple, is a common practice for feature combination \cite{zhou2015simple, huang2017densely}.

It is worth noting that we have described how to combine three encoders, \ie mean pooling, biGRU and biGRU-CNN, which naturally capture global, temporal and local patterns. In principle, other encoders, \eg Transformers, that are complementary to the existing encoders can be combined.

\subsection{Text-side Multi-level Encoding}

The above encoding network, after minor modification, is also applicable for the text modality. 

Given a sentence $s$ of length $m$, we represent each of its words by a one-hot vector. Accordingly, a sequence of one-hot vectors $\{w_1, w_2,\ldots, w_m\}$ is generated, where $w_t$ indicates the vector of the $t$-th word. Global encoding $f_s^{1}$ is obtained by averaging all the individual vectors in the sequence. This amounts to the classical bag-of-words representation. 

For biGRU based encoding, each word is first converted to a dense vector by multiplying its one-hot vector with a word embedding matrix. We initialize the matrix using a word2vec \cite{word2vec} model provided by \cite{dong2018predicting}, which trained word2vec on English tags of 30 million Flickr images. The rest is mostly identical to the video counterpart. We denote the biGRU based encoding of the sentence as $f_s^{2}$. Similarly, we have the biGRU-CNN based encoding of the sentence as $f_s^{3}$. 
Here, we utilize three 1D convolutional blocks with $k=2,3,4.$
Multi-level encoding of the sentence is obtained by concatenating the encoding results from all the three levels in the dual network, \ie
\begin{equation}
\phi(s) = [f_s^{1}, f_s^{2}, f_s^{3}].
\end{equation}

As $\phi(v)$ and $\phi(s)$ have not been correlated, they are not directly comparable. For video-text similarity computation, the vectors need to be projected into a common space, the learning algorithm for which will be presented next.

\section{Hybrid Space Learning}\label{ssec:learning}

We propose to train our dual encoding network with a hybrid space learning algorithm. The hybrid space consists of a latent space which aims for good performance and a concept space which is meant for good interpretability.

\subsection{Learning a Latent Space}

\textbf{Network}. 
Given the encoded video vector $\phi(v)$ and the sentence vector $\phi(s)$, we project them into a latent space by affine transformations. From the neural network viewpoint, an affine transformation is essentially a Fully Connected (FC) layer. 
We additionally use a Batch Normalization (BN) layer after the FC layer, as we find this trick beneficial. Putting everything together, we obtain the video feature vector $f(v)$ and sentence feature vector $f(s)$ in the latent space as:
\begin{equation} \label{eq:latent-network}
\begin{array}{l}
f(v) = \mbox{BN}(W_{1} \phi(v) + b_{1}), \\
f(s) = \mbox{BN}(W_{2} \phi(s) + b_{2}), \\
\end{array} 
\end{equation}
where $W_1$ and $ W_2$ parameterize the FC layers on each side, with $b_1$ and $b_2$ as  bias terms.

To measure the video-text similarity $sim_{lat}(v,s)$ in the latent space, we use popular cosine similarity between $f(v)$ and $f(s)$:  
\begin{equation}
sim_{lat}(v,s) = \frac{f(v)\cdot f(s)}{\left \| f(v) \right \| \left \| f(s) \right \|}.
\end{equation}
In our preliminary experiment, we also tried the Manhattan and Euclidean distance, but found them less effective than the cosine similarity.

\textbf{Loss}. A desirable similarity function shall make relevant video-sentence pairs near and irrelevant pairs far away in the latent space. Therefore, we use the improved triplet ranking loss \cite{faghri2017vse}, which penalizes the model according to the hardest negative examples in the mini-batch.
Concretely, given a relevant video-sentence pair $(v,s)$ in a mini-batch, its loss $\mathcal{L}_{lat}(v,s)$ is:
\begin{equation}\label{eq:l_loss}
\begin{array}{r}
\mathcal{L}_{lat}(v,s) = max(0, m + sim_{lat}(v,s^-) - sim_{lat}(v,s)) \\
 + max(0, m + sim_{lat}(v^-,s) - sim_{lat}(v,s)),
 \end{array} 
\end{equation}
where $m$ is the margin constant, while $s^-$ and $v^-$ respectively indicate a negative sentence sample for $v$ and a negative video sample for $s$. The two negatives are not randomly sampled. Instead, the most similar yet negative sentence and video in the current mini-batch are chosen.

\subsection{Learning a Concept Space}

As multiple concepts can be used simultaneously to describe a specific video or sentence, learning a concept space can be naturally formulated as a multi-label classification problem. 

\textbf{Network}. Suppose the size of the concept vocabulary is $K$. In order to project $\phi(v)$ and $\phi(s)$ into a $K$-dimensional concept space, we adopt a network similar to the network used for latent space learning. That is,
\begin{equation} \label{eq:concept-network}
\begin{array}{l}
g(v) = \sigma(BN(W_{3} \phi(v) + b_3)), \\
g(s) = \sigma(BN(W_{4} \phi(s) + b_4)), \\
\end{array} 
\end{equation}
Note that different from Eq. \ref{eq:latent-network}, we additionally use a sigmoid activation $\sigma$ to produce probabilistic output.
For a specific concept indexed by $i=1,\ldots,K$, we use $g(v)_i$ to denote the probability of the concept being relevant with respect to the video $v$. In a similar vein we define $g(s)_i$.

Consider the absolute scale of elements in the concept vectors matters, cosine similarity which mainly considers the direction of feature vectors is suboptimal to measure similarities between concept vectors.
Viewing the two concept vectors $g(v)$ and $g(s)$ as (unnormalized) histograms, we use generalized Jaccard similarity to compute the video-text similarity $sim_{con}(v,s)$ in the concept space, \ie
\begin{equation}
sim_{con}(v, s) = \frac{\sum_{i=1}^{K} \min(g(v)_i,g(s)_i)}{\sum_{i=1}^{K} \max(g(v)_i,g(s)_i)}.
\end{equation}

\begin{figure}[tb!]
\centering
\includegraphics[width=\columnwidth]{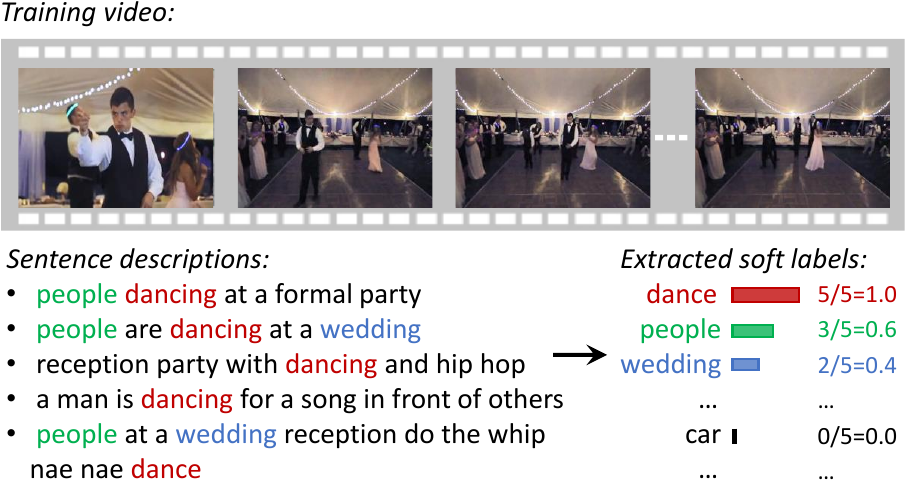}
\caption{\textbf{An illustration of extracting concept-level annotations from sentence descriptions of training videos}. Instead of binary labels, we extract frequency-based soft labels to better reflect the importance of a specific concept for a given video. E.g., the concept \emph{dance} appears in all the five sentences of the example video, so its $y$ value is $1$, whilst the concept \emph{wedding} occurring twice has a $y$ value of $0.4$}.
\label{fig:soft_label}
\end{figure}

\textbf{Concept-level annotations}. 
Per training video, we extract automatically its concept-level annotations from the associated sentence descriptions as follows.
Assume that for a specific training video $v$, we have access to $p$ sentences, $\{s_1,\ldots,s_p\}$, that describe the video content. 
The relevance of a specific concept w.r.t $v$ is determined by its occurrence in the $p$ sentences. Li \etal \cite{li2017measuring} suggest that a concept appearing in multiple sentences is usually more important than those presented once. Hence, instead of binary labels, we obtain soft labels based on concept frequency. Specifically, let $y$ be a $K$-dimensional ground-truth vector for $v$. The value of its $i$-th dimension, \ie $y_i$, is defined as the frequency of the $i$-th concept divided by the maximum frequency of all concepts within the $p$ sentences, see Fig. \ref{fig:soft_label}. 
Accordingly, we extend a relevant video-sentence pair $(v,s)$ to a triplet training instance $(v,s,y)$ to supervise concept space learning.

\textbf{Loss}. For multi-label classification, the binary cross-entropy (BCE) loss is common. In our context, the loss for a given video-sentence pair $(v,s)$ with respect to their shared ground-truth $y$ is computed as 
\begin{small}
\begin{equation}
\begin{split}
\mathcal{L}_{bce}(v,s,y) = - (\frac{1}{K}\sum_{i=1}^{K}[y_{i} \log(g(v)_{i})+(1-y_{i}) \log(1-g(v)_{i})] \\
 + \frac{1}{K}\sum_{i=1}^{K}[y_{i} \log(g(s)_{i})+(1-y_{i}) \log(1-g(s)_{i})]).
\end{split}
\end{equation}
\end{small}

We expect the concept space to be used not only for interpretability but also for improving video-text matching. So in addition to the BCE loss, we also minimize the improved triplet ranking loss in the concept space. That is, 
\begin{equation}\label{eq:p_triplet}
\begin{array}{r}
\mathcal{L}_{con,rank}(v,s) = max(0, m + sim_{con}(v,s^-) - sim_{con}(v,s)) \\
 + max(0, m + sim_{con}(v^-,s) - sim_{con}(v,s)).
 \end{array} 
\end{equation}
The concept space is learned by minimizing the following combined loss:
\begin{equation}\label{eq:p_loss}
\mathcal{L}_{con}(v,s,y) =  \mathcal{L}_{bce}(v,s,y) + \mathcal{L}_{con,rank}(v,s).
\end{equation}
Both $\mathcal{L}_{bce}$ and $\mathcal{L}_{con,rank}$ matter in Eq. \ref{eq:p_loss}. Without $\mathcal{L}_{bce}$, Eq. \ref{eq:p_loss} is boiled down to learning another latent space that not only lacks interpretability but also makes $sim_{con}(v,c)$ less complementary to $sim_{lat}(v,c)$ for video-text similarity computation. Meanwhile, using $\mathcal{L}_{bce}$ alone makes the concept space suboptimal for video-text matching and sensitive to the concept annotation quality, c.f. the appendix.

\subsection{Joint Learning of the Two Spaces}

The dual encoding network is trained by minimizing the combination of the latent-space loss $\mathcal{L}_{lat}$ and the concept-based loss $\mathcal{L}_{con}$.
In particular, given a training set $\mathcal{D}=\{(v,s,y)\}$, we have
\begin{equation}
\argmin_{\theta} \sum_{(v,s,y) \in \mathcal{D} } \mathcal{L}_{lat}(v,s) + \mathcal{L}_{con}(v,s,y),
\end{equation}
where $\theta$ denotes all the trainable parameters in the whole model.
Except for image CNNs used for video feature extraction, the dual encoding network is trained in an end-to-end manner.

\begin{figure}[tb!]
\centering\includegraphics[width=0.99\columnwidth]{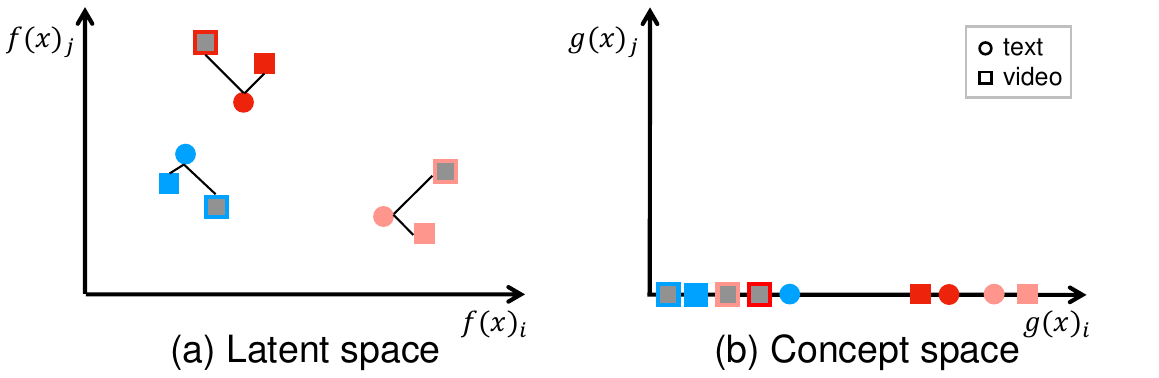}
\caption{\textbf{A toy example showing potential weakness of a latent space and a concept space in two dimensions}. Markers with the same color are relevant, while negative samples are with gray background. Red and light-red are semantically closer than the blue. Given a specific triplet, only their relative positions matter in the latent space. So even with zero triplet ranking loss, the red triplet can be projected into a region that is closer to the blue triplet rather than the light-red triplet. In the concept space, wherein each dimension corresponds to a distinct concept, the absolute positions of the triplets now matter. Samples positive w.r.t a specific concept are to be projected to one end, while negatives are to be placed near the origin. As the region around the origin naturally lacks discriminability, other dimensions have to be considered. However, as the concepts are pre-specified, they are not necessarily optimal to describe the video content. Best viewed in color.}
\label{fig:space_example}
\end{figure}

In order to conceptually explain why hybrid space learning helps, we provide a toy example in Fig. \ref{fig:space_example} that shows potential weakness when using a latent space or a concept space alone. As the latent space considers only relative positions per triplet, triplets that are semantically close, \ie red and light-red triplets in Fig. \ref{fig:space_example}(a), could be distant from each other. By contrast, each dimension of the concept space, which corresponds to a unique concept by definition, acts as an anchor to pinpoint the absolute position of a sample w.r.t. the concept. Fig. \ref{fig:space_example}(b) shows the embedding space w.r.t concept $i$ and concept $j$, where the concept $i$ alone is insufficient to discern samples nears its origin. Other dimensions have to be taken into account.  Nonetheless, as the concept set is pre-specified, the other dimensions are not necessarily discriminative, see the vertical axis corresponding to concept $j$.
The hybrid space learning gives us the possibility of combining the best of the two spaces, while overcoming their weaknesses. In addition, the combination of video-text similarities in the two distinct spaces can be viewed as a lightweight instantiation of model ensemble~\cite{kittler1998combining}, which typically improves over the base models.

\subsection{Video-Text Similarity Computation}
Once the model is trained, the final similarity between a video $v$ and a sentence $s$ is computed as the sum of their latent-space similarity and concept-space similarity, namely
\begin{equation}\label{eq:final_simi}
sim(v,s) = \alpha \cdot sim_{lat}(v,c) + (1-\alpha) \cdot sim_{con}(v,c),
\end{equation}
where $\alpha$ is a hyper-parameter to balance the importance of two spaces, ranging within $[0,1]$.
Note that raw values of $sim_{lat}(v,s)$ and $sim_{con}(v,s)$ reside in distinct scales. Hence, they are rescaled separately by min-max normalization before being combined. Also note that in the inference stage, the multi-level encoding at the video side can be performed independently. Hence, for a large-scale video collection, their hybrid-space features can be pre-computed, allowing us to answer ad-hoc queries on the fly.

\section{Evaluation} \label{sec:eval}

Our evaluation is organized as follows. Firstly in Section \ref{ssec:exp-sota}, we compare the proposed Dual Encoding model (with its best setup) against the state-of-the-art on four datasets, \ie MSR-VTT~\cite{xu2016msr}, TRECVID AVS 2016-2018~\cite{AwadTRECVID16,AwadTRECVID17,AwadTRECVID18}, VATEX~\cite{wang2019vatex} and MPII-MD~\cite{rohrbach2015dataset}. Constructed independently by the dataset developers, the first three datasets consist of short web videos with very diverse content, while the last dataset contains video clips from 72 movies. 
Second, in order to verify the influence of major components in the proposed model, Section \ref{ssec:ablation} presents an ablation study on the MSR-VTT dataset.  Additional experiments concerning the effect of the hyper-parameter $\alpha$ and the concept annotation quality on the performance and the efficiency of our model are provided in the appendix.

Before proceeding to the experiments, we detail common implementations regarding text preprocessing, video features, concept vocabulary extraction, and model training. 
For sentence preprocessing, we first convert all words to the lowercase and then replace words that occurring less than five times in the training set with a special token. 
For video features, on VATEX we adopt 1,024-d I3D \cite{carreira2017quo} video features provided by the dataset developers~\cite{wang2019vatex}.
As for the other datasets, we extract frame-level ResNeXt-101~\cite{xie2017aggregated,mettes2020shuffled} and ResNet-152~\cite{cvpr2016-resnet} using an open-source toolbox\footnote{\url{https://github.com/xuchaoxi/video-cnn-feat}}. The two feature vectors are concatenated to obtain a combined 4,096-d CNN feature, which we refer to as ResNeXt-ResNet.

To obtain the concept vocabulary, we conduct part-of-speech tagging by NLTK toolkit on all training sentences, and only keep the nouns, verbs and adjectives. All the English stopwords also removed. Besides, we also lemmatize the words, making \textit{dog} and \textit{dogs} to be a same concept. Finally, the top $K=512$ frequent words are selected as the final concept vocabulary.

The proposed model is implemented using PyTorch (http://pytorch.org). Following~\cite{faghri2017vse}, the parameter $m$ for the improved triplet ranking loss is set to $0.2$. The weight $\alpha$ in the combined similarity is empirically set to $0.6$. 
We learn from our earlier studies~\cite{li2019w2vv++,cvpr2019-dual-dong} that setting the dimensionality of the common space to $2,048$ is a good practice. Hence, we let the overall dimensionality of the hybrid space be $2,048$. Recall that the concept space is 512-dimensional. Accordingly, the dimensionality of the latent space is $2,048-512=1,536$.
We use stochastic gradient descent with Adam \cite{kingma2014adam}. The mini-batch size is 128. With an initial learning rate of 0.0001, we take an adjustment schedule similar to \cite{dong2018predicting}. That is, once the validation loss does not decrease in three consecutive epochs, we divide the learning rate by 2. Early stop occurs if the validation performance does not improve in ten consecutive epochs. The maximal number of epochs is 50. In practice, early stop occurs typically after 15 epochs.

\subsection{Comparison with the State-of-the-art} \label{ssec:exp-sota}

\subsubsection{Experiments on MSR-VTT}

\textbf{Data}. 
The MSR-VTT dataset \cite{xu2016msr}, originally developed for video captioning, consists of 10k web video clips and 200k natural sentences describing the visual content of the clips. The number of sentences per clip is 20. 
For this dataset, we notice there are three distinct editions of data partition in the literature~\cite{xu2016msr,miech2018learning,yu2018joint}.
The official partition~\cite{xu2016msr} uses 6,513 clips for training, 497 clips for validation, and the remaining 2,990 clips for testing.
For the partition by \cite{miech2018learning}, there are 6,656 clips for training and 1,000 clips for testing. The partition of \cite{yu2018joint} uses 7,010 and 1,000 clips for training and testing, respectively. 
As the last two data partitions provide no validation set, we build a validation set by randomly sample 1,000 clips from MSR-VTT with \cite{miech2018learning,yu2018joint} excluded, respectively.
For a comprehensive evaluation, our experiments are performed on all the three data partitions.

\textbf{Performance Metrics}.  We use rank-based metrics, namely $R@K$ ($K = 1, 5, 10$), Median rank (Med r) and mean Average Precision (mAP) to evaluate the performance. $R@K$ is the percentage of test queries for which at least one relevant item is found among the top-$K$ retrieved results. Med r is the median rank of the first relevant item in the search results. Higher $R@K$, mAP and lower Med r mean better performance. 
For overall comparison, we report the Sum of all Recalls (SumR).

\begin{table*} [tb!]
\renewcommand{\arraystretch}{1}
\caption{\textbf{State-of-the-art on MSR-VTT}. Larger R@\{1,5,10\}, mAP and smaller Med r indicate better performance. Symbol asterisk (*) indicates numbers directly cited from the original papers, and the others are obtained by our re-training given the same ResNeXt-ResNet feature. On all the three distinct editions of data partition, the proposed Dual Encoding model obtains the best overall performance.
}
\label{tab:sota-msrvtt}
\centering 
\scalebox{0.93}{
\begin{tabular}{@{}l*{12}{r}c @{}}
\toprule
\multirow{2}{*}{\textbf{Method}}   & \multicolumn{5}{c}{\textbf{Text-to-Video Retrieval}} && \multicolumn{5}{c}{\textbf{Video-to-Text Retrieval}} & \multirow{2}{*}{\textbf{SumR}} \\
 \cmidrule{2-6}  \cmidrule{8-12} 
& R@1 & R@5 & R@10 & Med r & mAP && R@1 & R@5 & R@10 & Med r & mAP & \\
\cmidrule{1-13}
\textbf{Official full-size test set \cite{xu2016msr}}  \\
Francis \etal* \cite{iccv2019-francis}     & 6.5 & 19.3 & 28.0 & 42 & - && - & - & - & - & - & -  \\
Mithun \etal* \cite{mithun2018learning}    & 7.0 & 20.9 & 29.7 & 38 & - &&    12.5 & 32.1 & 42.4 & 16 & - &     144.6 \\
TCE* \cite{sigir2020tree}                  & 7.7 & 22.5 & 32.1 & 30 & - && - & - & - & - & - & -  \\
HGR* \cite{chen2020fine}            & 9.2  & 26.2 & 36.5 & 24 & - &&          15.0 & 36.7 & 48.8 & 11 & - &     172.4 \\
CE* \cite{liu2019use}               & 10.0 & 29.0  & 41.2  & \textbf{16}  & -   & &    15.6  & 40.9  & 55.2   & 8.3  & - & 191.9 \\
\cmidrule{2-13}
W2VV \cite{dong2018predicting}    & 1.1  & 4.7  & 8.1 & 236 & 3.7 &&      17.0 & 37.9 & 49.1 & 11 & 7.6 &     117.9 \\
MEE \cite{miech2018learning}      & 6.8 & 20.7 & 31.1 & 28 & 14.7  &&   13.4 & 32.0 & 44.0 & 14 & 6.6  & 148.0 \\
CE \cite{liu2019use}              & 7.9 & 23.6 & 34.6 & 23 & 16.5 &&       11.0 & 31.9 & 46.1 & 13  & 6.8 & 155.1 \\
VSE++ \cite{faghri2017vse}        & 8.7  & 24.3 & 34.1 & 28 & 16.9 &&      15.6 & 36.6 & 48.6 & 11 & 7.4 & 167.9 \\
TCE \cite{sigir2020tree}          & 9.3 & 27.3 & 38.6 & 19 & 18.7 &&       15.1 & 36.8 & 50.2 & 10 & 8.0 & 177.3 \\
W2VV++ \cite{li2019w2vv++}        & 11.1 & 29.6 & 40.5 & 18 & 20.6 &&      17.5 & 40.2 & 52.5 & 9 & 8.5 & 191.4 \\
HGR \cite{chen2020fine}           & 11.1 &	\textbf{30.5} &	\textbf{42.1} & \textbf{16} & 20.8 &&    18.7 & 44.3 & 57.6 & \textbf{7} & 9.9 & 204.4 \\
\textit{Dual Encoding}         & \textbf{11.6} & 30.3 & 41.3 & 17 & \textbf{21.2} && \textbf{22.5} & \textbf{47.1} & \textbf{58.9} & \textbf{7} & \textbf{10.5} & \textbf{211.7} \\  [3pt]

\hline
\textbf{Test1k-Miech \cite{miech2018learning}}  \\
JPoSE*  \cite{wray2019fine}                & 14.3 & 38.1 & 53.0 & 9 & - && 16.4 & 41.3 & 54.4 & 8.7 & - & 217.5 \\
MEE* \cite{miech2018learning}               & 16.8 & 41.0 & 54.4 & 9 & - && - & - & - & - & - & -  \\
TCE* \cite{sigir2020tree}                  & 17.1 & 39.9 & 53.7 & 9 & - && - & - & - & - & - & -  \\
CE* \cite{liu2019use}                      & 18.2 & 46.0 & 60.7 & 7 & - && 18.0 & 46.0  & 60.3 & 6.5 & - & 249.2 \\
\cmidrule{2-13}
W2VV \cite{dong2018predicting}            & 2.7 & 12.5 & 17.3 & 83 & 7.9 &&      17.3 & 42.0 & 53.5 & 9 & 29.3 & 145.3 \\
MEE \cite{miech2018learning}             & 15.7 & 39.0 & 52.3 & 9 & 27.1  && 15.3 & 41.9 & 54.5& 8 & 28.1 & 218.7 \\
VSE++ \cite{faghri2017vse}               & 17.0 & 40.9 & 52.0 & 10 & 16.9 &&     18.1 & 40.4 & 52.1 & 9 & 29.2 & 220.5 \\
CE \cite{liu2019use}                     & 17.8 & 42.8 & 56.1 & 8 & 30.3  &&   17.4 & 42.9 & 56.1 & 8 & 29.8 & 233.1 \\
TCE \cite{sigir2020tree}                 & 17.0 & 44.7 & 58.3 & 7 & 30.0  &&   15.1 & 43.3 & 58.2 & 7 & 28.3 & 236.6 \\
W2VV++ \cite{li2019w2vv++}               & 21.7 & 48.6 & 60.9 & 6 & 34.4 &&    18.6 & 46.4 & 59.1 & 6 & 31.7 & 255.3 \\
HGR \cite{chen2020fine}                  & 22.9 & 50.2 & \textbf{63.6} & \textbf{5} & 35.9  &&   20.0 & 48.3 & 60.9 & 6 & 33.2 & 265.9 \\
\textit{Dual Encoding}         & \textbf{23.0} & \textbf{50.6} & 62.5 & \textbf{5} & \textbf{36.1} && \textbf{25.1} & \textbf{52.1} & \textbf{64.6} & \textbf{5} & \textbf{37.7} & \textbf{277.9} \\ [3pt]

\hline
\textbf{Test1k-Yu \cite{yu2018joint}} \\
CT-SAN*   \cite{yu2017end}                 & 4.4 & 16.6 & 22.3 & 35 & - && - & - & - & - & - & -  \\
JSFusion* \cite{yu2018joint}               & 10.2 & 31.2 & 43.2 & 13 & - && - & - & - & - & - & -  \\
TCE* \cite{sigir2020tree}                  & 16.1 & 38.0 & 51.5 & 10 & - && - & - & - & - & - & -  \\
Miech \etal*  \cite{miech2019howto100m}    & 14.9 & 40.2 & 52.8 & 9 & - && - & - & - & - & - & - \\
CE* \cite{liu2019use}                      & 20.9  & \textbf{48.8}  & \textbf{62.4}  & \textbf{6} & - && 20.6  & \textbf{50.3}  & \textbf{64.0}  & \textbf{5.3}  & - & \textbf{267.0} \\
\cmidrule{2-13}
W2VV \cite{dong2018predicting}            & 1.9 & 9.9 & 15.2 & 79 & 6.8 &&    17.3 & 39.3 & 50.2 & 10 & 27.8 & 133.8 \\
VSE++ \cite{faghri2017vse}               & 16.0 & 38.5 & 50.9 & 10 & 27.4 &&      16.2 & 39.3 & 51.2 & 10 & 27.4 & 212.1 \\
MEE \cite{miech2018learning}             & 14.6 & 38.4 & 52.4 & 9 & 26.1 && 15.2 & 40.9 & 53.8 & 9 & 27.9 & 215.3 \\
W2VV++ \cite{li2019w2vv++}                & 19.0 & 45.0 & 58.7 & 7 & 31.8 &&   16.9 & 42.7 & 54.6 & 8 & 29.0 & 236.9 \\
CE \cite{liu2019use}                     & 17.2	& 46.2 & 58.5 & 7  & 30.3 && 15.8 & 44.9 & 59.2 & 7 & 30.4 & 241.8 \\
TCE \cite{sigir2020tree}                 & 17.8 & 46.0 & 58.3 & 7 & 31.1 && 18.9 & 43.5 & 58.8 & 7 & 31.4 & 243.3 \\
HGR \cite{chen2020fine}                  & \textbf{21.7} & 47.4 & 61.1 & 6 & \textbf{34.0} && 20.4 & 47.9 & 60.6 & 6 & 33.4 & 259.1 \\
\textit{Dual Encoding}                      & 21.1 & 48.7 & 60.2 & \textbf{6} & 33.6 &&      \textbf{21.7} & 49.4 & 61.6 & 6 & \textbf{34.7} & 262.7 \\
\bottomrule
\end{tabular}
 }
\end{table*}

\textbf{Baselines}. The following thirteen state-of-the-art models are compared:\\
$\bullet$ VSE++ \cite{faghri2017vse}: A state-of-the-art text-image retrieval model, which is commonly used as the strong baseline model for text-video retrieval. We replace its image-side branch with mean pooling on frame-level feature followed by a FC layer. \\ 
$\bullet$ W2VV~\cite{dong2018predicting}: Learn to project text into a visual feature space by minimizing the distance of relevant video-text pairs in the visual space. Multiple text encoding strategies including BoW, word2vec and GRU are jointly used for text encoding.  \\
$\bullet$ MEE~\cite{miech2018learning}: Use four different features to represent videos, and learn one latent space for each video feature. The weighted sum of similarities in four latent spaces is regarded as the final video-text similarity. \\
$\bullet$ W2VV++~\cite{li2019w2vv++}: An improved version of W2VV, it employs a better sentence encoding strategy and an improved triplet ranking loss.  \\
$\bullet$ CE~\cite{liu2019use}: Use a collaborative gating to fuse multiple features to obtain a strong video representation. \\
$\bullet$ TCE \cite{sigir2020tree}: Utilize a latent semantic tree \cite{tai2015improved} augmented encoder to represent text, and a GRU with multi-head self-attention mechanism \cite{vaswani2017attention} to encode videos. \\
$\bullet$ HGR \cite{chen2020fine}: Utilize graph convolutional network to model the connection between words, and project text and videos into three latent spaces. \\
$\bullet$ Mithun \etal \cite{mithun2018learning}: Project videos and text into two latent spaces, and a weighted triplet ranking loss is used for training.  \\
$\bullet$ Francis \etal \cite{iccv2019-francis}: Fuse multimodal features, \ie counting, activity and concpet features to obtain stronger video and text representations. \\
$\bullet$ JPoSE \cite{wray2019fine}: Decompose text into nouns and non-noun words, and respectively project them into two different latent spaces. \\
$\bullet$ CT-SAN \cite{yu2017end}: Learn to directly predict the similarity based on the fused video-text features without learning a common space, and a concept word detector is used for enhancing the video representation.    \\
$\bullet$ JSFusion \cite{yu2018joint}: With the same idea of \cite{yu2017end} that direly predicts the video-text similarity, a stronger joint sequence fusion is employed to fuse video and text features. \\
$\bullet$ Miech \etal \cite{miech2019howto100m}: Project videos and text into a common space by a gated embedding module respectively. The model is pre-trained on large-scale video-text dataset HowTo100M \cite{miech2019howto100m} and fine-tuned on MSR-VTT.

For a direct comparison, we cite numbers from the original papers whenever applicable. Meanwhile, we notice that video features used by specific papers vary. So, to make the comparison fairer, we have re-trained the following seven models which have been open-sourced, \ie W2VV, MEE, VSE++, W2VV++, CE, TCE and HGR, using the same ResNeXt-ResNet feature\footnote{MEE and CE employ a separated branch to handle the two video features respectively, while others utilize the concatenated feature as the whole input. }.

\textbf{Results}.
Table \ref{tab:sota-msrvtt} summarizes the performance comparison on three different data partitions of MSR-VTT. Though our goal is video retrieval by text, which corresponds to text-to-video retrieval in the table, video-to-text retrieval is also included for completeness.
Note that the number of the candidate videos/sentences to be retrieved in the official partition is larger than that in the other two partitions. Hence, the official partition is more challenging. As a consequence, for all the models, their performance scores on the official partition are lower than their counterparts on the other partitions. 
Consider the results using the same video features, our dual encoding model achieves the best overall performance.

Among the results marked with asterisks (*), CE* which utilizes seven video features performs the best.
Still, the proposed Dual Encoding model using only two visual features outperforms CE* on the first two data partitions, while slightly worse on Test1k-Yu. Note that on this test set, Dual Encoding was trained on the corresponding training set of 7,010 videos as specified by \cite{yu2018joint}, whilst CE* was trained on an enlarged set of 9,000 videos. The results demonstrate the effectiveness of our proposed model.

\subsubsection{Experiments on TRECVID AVS 2016-2018}

\textbf{Data}. 
IACC.3 dataset is the largest test bed for video retrieval by text to this date, which developed for TRECVID (Ad-hoc Video Search) AVS 2016, 2017 and 2018 task \cite{AwadTRECVID16,AwadTRECVID17,AwadTRECVID18}. 
The dataset contains 4,593 Internet Archive videos with duration ranging from 6.5 minutes to 9.5 minutes and a mean duration of almost 7.8 minutes. Shot boundary detection results in 335,944 shots in total. 
Given an ad-hoc query, \eg \emph{Find shots of military personnel interacting with protesters}, the task is to return for the query a list of 1,000 shots from the test collection ranked according to their likelihood of containing the given query. Per year TRECVID specifies 30 distinct queries of varied complexity.

As TRECVID does not specify training data for the AVS task, we train the dual encoding network using the joint collection of MSR-VTT and the TGIF \cite{tgif} which contains 100K animated GIFs and 120K sentences describing visual content of the GIFs. Although animated GIFs are a very different domain, TGIF was constructed in a way to resemble user-generated video clips, \eg with cartoon, static, and textual content removed. For IACC.3, MSR-VTT and TGIF, we use the ResNeXt-ResNet video feature.

\textbf{Performance Metrics}. 
We utilize inferred Average Precision (infAP), the official performance metric used by the TRECVID AVS task. The overall performance is measured by averaging infAP scores over the queries. Their values are reported in percentage (\%).
Note that the TRECVID ground truth is partially available at the shot-level. The task organizers employ a pooling strategy to collect the ground truth, \ie a pool of candidate shots are formed by collecting the top-1000 shots from each submission and a random subset is selected for manual verification. The ground truth thus favors official participants.
As the top ranked items found by our method can be outside of the subset, infAP scores of our method are likely to be underestimated.

\textbf{Baselines}.
We include the top 3 entries of each year, \ie \cite{tv16-nii,tv16-certh,tv16-inf} for 2016, \cite{tv17-uva,tv17-waseda,tv17-vireo} for 2017 and \cite{li2018renmin,tv18-informedia,tv18-ntu} for 2018. Besides we include publications on the tasks, \ie \cite{pami2017-videostory,icmr2017-certh-avs}. 
The most of above methods are concept based, except \cite{li2018renmin,tv18-informedia,tv18-ntu}. 
Among them, \cite{li2018renmin} fuses three W2VV++ variants with different settings, 
\cite{tv18-informedia} uses two attention networks, besides the classical concept-based representation, while \cite{tv18-ntu} is based on VSE++.
Notice that visual features and training data used by these methods vary, meaning the comparison and consequently conclusions drawn from this comparison is at a system level. So for a more conclusive comparison, we re-train VSE++~\cite{faghri2017vse}, W2VV~\cite{dong2018predicting}, W2VV++~\cite{li2019w2vv++} and CE~\cite{liu2019use} using the same training data and the ResNeXt-ResNet feature.

\textbf{Results}.
Table \ref{tab:avs-perf} shows the performance of different methods on the TRECVID AVS 2016, 2017 and 2018 tasks, and the overall performance is the mean score of the three years. The proposed method again performs the best, with infAP of 15.2, 23.1 and 12.1 respectively. While \cite{li2018renmin} has a same infAP of 12.1 on the TRECVID AVS 2018 task, their solution ensembles three models. Their best single model, \ie W2VV++ \cite{li2019w2vv++}  which uses the same training data and the same ResNeXt-ResNet feature, has a lower infAP of 10.6. 
Given the same training data and feature, the proposed method outperforms VSE++, W2VV, CE with a clear margin. 
These results confirm the effectiveness of our dual encoding for large-scale video retrieval by text.

\begin{table} [tb!]
\renewcommand{\arraystretch}{1.2}
\caption{\textbf{State-of-the-art on the TRECVID AVS 2016 / 2017 / 2018}.
Symbol asterisk (*) indicates numbers directly cited from the original papers. The proposed dual encoding model consistently perform the best. 
}
\label{tab:avs-perf}
\centering
 \scalebox{0.93}{
 \begin{tabular}{@{} l lllr @{}}
\toprule
         & \multicolumn{3}{c}{\textbf{TRECVID edition}} \\
         \cmidrule{2-4}
         & \emph{2016} & \emph{2017} & \emph{2018} & \textbf{OVERALL} \\
        \midrule
\textbf{Top-3 TRECVID finalists}: \\        
Rank 1* & 5.4 \cite{tv16-nii}   & 20.6 \cite{tv17-uva}    & \textbf{12.1} \cite{li2018renmin} & -- \\
Rank 2* & 5.1 \cite{tv16-certh} & 15.9 \cite{tv17-waseda} & 8.7 \cite{tv18-informedia} & -- \\
Rank 3* & 4.0 \cite{tv16-inf}   & 12.0 \cite{tv17-vireo}  & 8.2 \cite{tv18-ntu} & -- \\
\cmidrule{1-1}
\textbf{Literature methods}: \\
VideoStory* \cite{pami2017-videostory,tv2017-uva-avs} &  8.7 & 15.0 & -- & -- \\
Markatopoulou \etal* \cite{icmr2017-certh-avs}        &  6.4 & -- & -- & -- \\
CE \cite{liu2019use}                                &  7.4 & 14.5 & 8.6 & 10.2 \\
VSE++ \cite{faghri2017vse}                          &  13.5 & 16.3 & 10.6 & 13.5 \\
W2VV \cite{dong2018predicting}                      &  14.9 & 19.8 & 10.3 & 15.0 \\  
W2VV++ \cite{li2019w2vv++}                          &  15.1 & 21.3 & 10.6 & 15.7 \\
\textit{Dual Encoding}      & \textbf{15.2} & \textbf{23.1} & \textbf{12.1} & \textbf{16.8}  \\
\bottomrule
\end{tabular}
 }
\end{table}

\subsubsection{Experiments on VATEX}

\begin{table} [tb!]
\renewcommand{\arraystretch}{1.2}
\caption{\textbf{State-of-the-art on VATEX}. Our proposed model performs the best.}
\label{tab:vatex_perf}
\centering 
\scalebox{0.93}{
\begin{tabular}{@{}l*{8}{r} @{}}
\toprule
\multirow{2}{*}{\textbf{Method}}   & \multicolumn{3}{c}{\textbf{Text-to-Video}} && \multicolumn{3}{c}{\textbf{Video-to-Text}} & \multirow{2}{*}{\textbf{SumR}} \\
 \cmidrule{2-4}  \cmidrule{6-8} 
& R@1 & R@5 & R@10 && R@1 & R@5 & R@10 &  \\
\cmidrule{1-9}
W2VV   \cite{dong2018predicting}     & 14.6 & 36.3 & 46.1   &&   39.6 & 69.5 & 79.4 & 285.5 \\
VSE++  \cite{faghri2017vse}          & 31.3 & 65.8 & 76.4   &&   42.9 & 73.9 & 83.6 & 373.9   \\
CE     \cite{liu2019use}             & 31.1 & 68.7 & 80.2   &&   41.3 & 71.0 & 82.3 & 374.6 \\
W2VV++ \cite{li2019w2vv++}           & 32.0 & 68.2 & 78.8   &&   41.8 & 75.1 & 84.3 & 380.2   \\
HGR    \cite{chen2020fine}           & 35.1 & 73.5 & 83.5   &&   - & - & - & - \\
\textit{Dual Encoding}  & \textbf{36.8} & \textbf{73.6} & \textbf{83.7}   &&   \textbf{46.8} & \textbf{75.7} & \textbf{85.1} & \textbf{401.7}  \\
\bottomrule
\end{tabular}
 }
\end{table}

\textbf{Data}. 
VATEX \cite{wang2019vatex} a large-scale multilingual video description dataset. Each video, collected for YouTube, has a duration of 10 seconds. Per video there are 10 English sentences and 10 Chinese sentences to describe the corresponding video content. Here, we only use the English sentences. 
We adopt the dataset partition provided by \cite{chen2020fine}, \ie 25,991 video clips for training, 1,500 clips for validation and 1,500 clips for testing, where validation and test set are obtained by randomly split the official validation set of 3,000 clips into two equal parts.

\textbf{Performance Metrics}. 
$R@K$ ($K = 1, 5, 10$) and SumR are used as the performance metrics.

\textbf{Baselines}. For method comparison, we consider HGR~\cite{chen2020fine}, the first work reporting video retrieval performance on VATEX. We also compare the VSE++~\cite{faghri2017vse}, W2VV~\cite{dong2018predicting}, W2VV++~\cite{li2019w2vv++} and CE~\cite{liu2019use}.

\textbf{Results}.
Table \ref{tab:vatex_perf} summarizes the performance, where all the models use the same I3D \cite{carreira2017quo} video features. Among them, VSE, VSE++ and W2VV++ are all use mean pooling over the frame-level feature to encode videos, while our dual encoding model explores multi-level features to represent videos, consistently achieving better performance. The result shows the benefit of using the multi-level feature for video representation.
Although HGR utilizes the extra semantic role annotation of sentences for text representation, our dual encoding model still slightly outperforms HGR.

\subsubsection{Experiments on MPII-MD}

\begin{table} [tb!]
\renewcommand{\arraystretch}{1.2}
\caption{\textbf{State-of-the-art on MPII-MD}. Our proposed model performs the best.}
\label{tab:mpii_perf}
\centering 
\scalebox{0.93}{
\begin{tabular}{@{}l*{8}{r} @{}}
\toprule
\multirow{2}{*}{\textbf{Method}}   & \multicolumn{3}{c}{\textbf{Text-to-Video}} && \multicolumn{3}{c}{\textbf{Video-to-Text}} & \multirow{2}{*}{\textbf{SumR}} \\
 \cmidrule{2-4}  \cmidrule{6-8} 
& R@1 & R@5 & R@10 && R@1 & R@5 & R@10  \\
\cmidrule{1-9}
W2VV++ \cite{li2019w2vv++}           & 0.3 & 1.0 & 1.7   &&   0.1 & 0.7 & 1.3 & 5.1 \\
VSE++  \cite{faghri2017vse}          & 0.2 & 0.9 & 1.6   &&   0.8 & 2.2 & 3.6 & 9.3 \\
W2VV   \cite{dong2018predicting}     & 0.1 & 0.3 & 0.5   &&   1.3 & 4.0 & 6.1 & 12.3 \\
CE     \cite{liu2019use}             & 0.9 & 3.1 & 5.7   &&   1.1 & 3.6 & 5.8 & 20.2 \\
\textit{Dual Encoding}  & \textbf{1.7} & \textbf{4.8} & \textbf{7.0}   &&   \textbf{1.4} & \textbf{4.7} & \textbf{7.0} & \textbf{26.6} \\
\bottomrule
\end{tabular}
 }
\end{table}

\begin{figure}[tb!]
\centering
\includegraphics[width=\columnwidth]{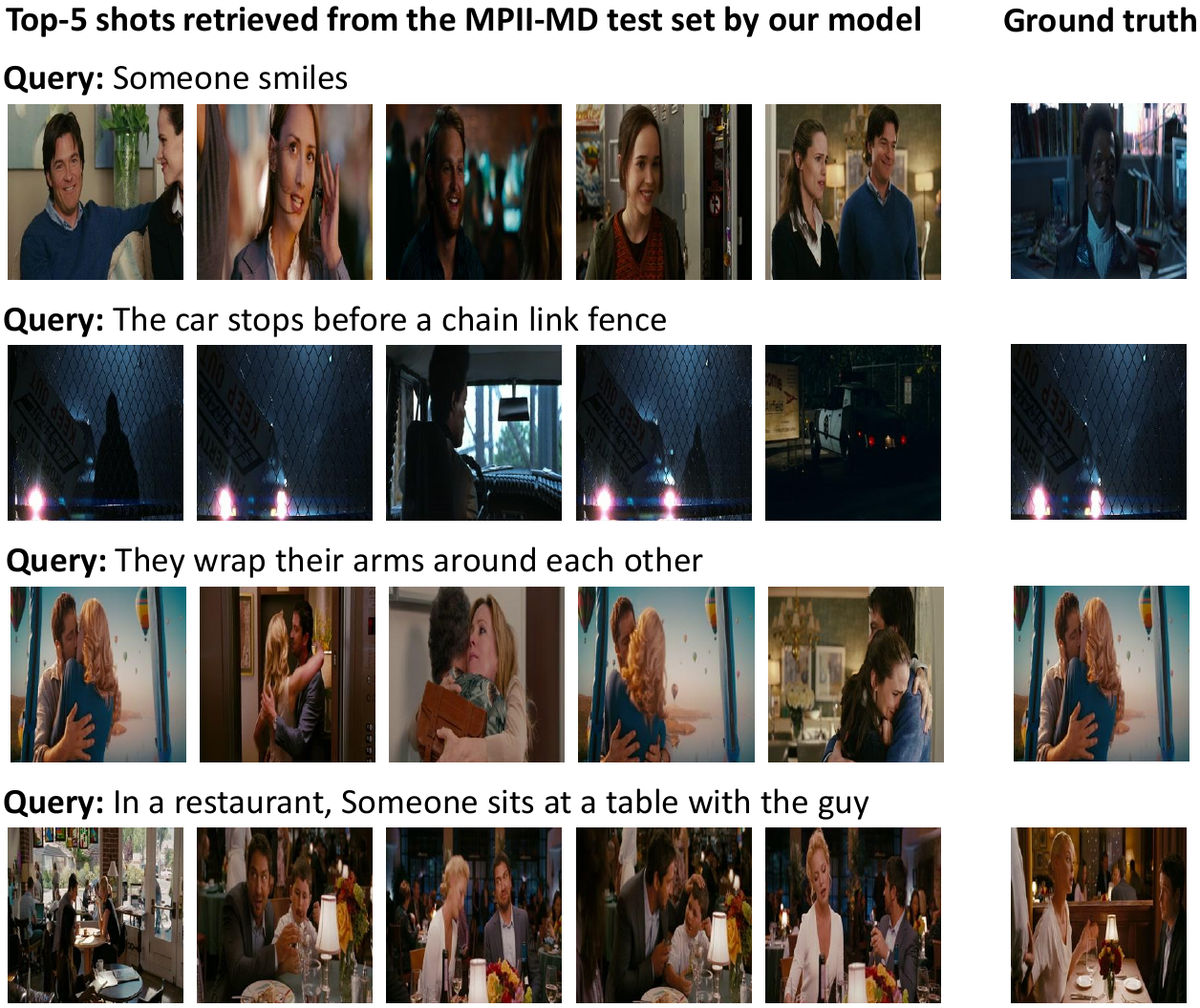}
\caption{\textbf{Selected examples of movie retrieval by text on MPII-MD}. The top retrieved shots, though not being ground truth, appear to be correct.}
\vspace{-4mm}
\label{fig:mpii-md}
\end{figure}

\begin{table*} [tb!]
\renewcommand{\arraystretch}{1.2}
\caption{\textbf{Effectiveness of Dual Encoding}. The overall performance, as indicated by Sum of Recalls, goes up as more encoding layers are added. Dual encoding exploiting all the three levels is the best.}
\label{tab:ablation-encoding}
\centering 
\scalebox{0.93}{
\begin{tabular}{l*{12}{r}c}
\toprule
\multirow{2}{*}{\textbf{Encoding strategy}} & \multicolumn{5}{c}{\textbf{Text-to-Video Retrieval}} && \multicolumn{5}{c}{\textbf{Video-to-Text Retrieval}} &  \multirow{2}{*}{\textbf{SumR}}\\
\cmidrule(l){2-6} \cmidrule(l){8-12}
  &  R@1 & R@5 & R@10  & Med r & mAP && R@1 & R@5 & R@10  & Med r & mAP & \\
\cmidrule(l){1-13}
Level 1 (Mean pooling)    &   9.7 & 26.8 & 37.0 & 23 & 18.5  &&     17.7 & 40.0 & 51.7 & 10 & 8.3 &     182.9 \\
Level 2 (biGRU)           &  10.3 & 28.5 & 39.4 & 19 & 19.7  &&     18.4 & 41.9 & 54.3 & 8  & 9.3 &     192.7 \\
Level 3 (biGRU-CNN)       &  11.1 & 30.1 & 41.6 & 17 & 20.9  &&     18.1 & 41.6 & 55.3 & 8  & 9.6 &     197.9 \\
Level 1 + 2               &  10.6 & 28.8 & 39.2 & 20 & 19.9  &&     19.1 & 43.1 & 54.5 & 8  & 9.2 &     195.3 \\
Level 1 + 3               &  11.5 & 30.0 & 40.8 & 18 & 20.9  &&     19.8 & 42.7 & 55.2 & 8  & 9.5 &     200.1 \\
Level 2 + 3               &  11.4 & \textbf{30.6} & \textbf{41.7} & \textbf{17} & \textbf{21.2} &&      19.9 & 44.3 & 55.8 & 8  & 10.1 &     203.8 \\ 
Level 1 + 2 + 3           & \textbf{11.6} & 30.3 & 41.3 & \textbf{17} & \textbf{21.2} && \textbf{22.5} & \textbf{47.1} & \textbf{58.9} & \textbf{7} & \textbf{10.5} & \textbf{211.7} \\ 
\bottomrule
\end{tabular}
 }
\end{table*}

\begin{table*} [tb!]
\renewcommand{\arraystretch}{1.2}
\caption{\textbf{Performance of Dual encoding with distinct common spaces}. Dataset: MSR-VTT.}
\label{tab:ablation-space}
\centering 
\scalebox{0.9}{
\begin{tabular}{l*{12}{r}c}
\toprule
\multirow{2}{*}{\textbf{Common Space}} & \multicolumn{5}{c}{\textbf{Text-to-Video Retrieval}} && \multicolumn{5}{c}{\textbf{Video-to-Text Retrieval}} &  \multirow{2}{*}{\textbf{SumR}}\\
\cmidrule(l){2-6} \cmidrule(l){8-12}
  &  R@1 & R@5 & R@10  & Med r & mAP && R@1 & R@5 & R@10  & Med r & mAP & \\
\cmidrule(l){1-13}
2,048-d Latent Space (Conference version~\cite{cvpr2019-dual-dong})  & 11.0 & 29.2 & 39.8 & 19 & 20.2  &&      18.8 & 42.7 & 56.2 & 8 & 9.3 &  197.7 \\
1,536-d Latent Space  & 11.0 & 29.3 & 39.9 & 19 & 20.3 && 19.7 & 43.6 & 55.6 & 8 & 9.3  & 199.0   \\
512-d Concept Space   & 9.9 & 26.8 & 37.4 & 23 & 18.7 && 17.9 & 41.5 & 53.9 & 8 & 9.0 &  187.4 \\
2,048-d Hybrid, without $\mathcal{L}_{con,rank}$ & 9.8 & 26.3 & 36.0 & 25 & 18.2 && 17.2 & 40.7 & 53.3 & 9 & 8.7 &  183.3 \\
2,048-d Hybrid: 1,536-d Latent Space + 512-d Concept Space       & \textbf{11.6} & \textbf{30.3} & \textbf{41.3} & \textbf{17} & \textbf{21.2} && \textbf{22.5} & \textbf{47.1} & \textbf{58.9} & \textbf{7} & \textbf{10.5} & \textbf{211.7} \\ 
\bottomrule
\end{tabular}
 }
\end{table*}

\textbf{Data}. 
To evaluate the effectiveness of our proposed methods for a specific video domain, we conduct the experiment on MPII-MD \cite{rohrbach2015dataset} a movie description dataset.
We use the official data partition, that is, 56,828, 4,929 and 6,580 movie clips for training, evaluation and testing, respectively. 
Each movie clip is associated with one or two textual descriptions.

\textbf{Performance Metrics}. 
Performance of $R@1$, $R@5$, $R@10$ and SumR are reported.

\textbf{Baselines}. In this experiments, we compare the VSE++~\cite{faghri2017vse}, W2VV~\cite{dong2018predicting}, W2VV++~\cite{li2019w2vv++} and CE~\cite{liu2019use}. All the models are trained using the  ResNeXt-ResNet feature.

\textbf{Results}. 
Table \ref{tab:mpii_perf} summarizes the performance on the MPII-MD dataset.
Our proposed dual encoding model outperforms the other counterparts. 
It is worth noting that the performance of all models are lower on MPII-MD than that on MSR-VTT, we attribute it to the challenging nature of movie retrieval by vague descriptions on MPII-MD.
See Fig.~\ref{fig:mpii-md} for some qualitative results.

\subsection{Ablation Study} \label{ssec:ablation}

In this section, we evaluate the viability of each component on MSR-VTT, with performance reported on the full-size test set unless otherwise stated. A comparison with recent multi-modal Transformer methods is also conducted. In addition, we investigate the effectiveness of our multi-level text encoding for image-text retrieval.

\subsubsection{Multi-level Encoding versus Single-Level Encoding}
To exam the usefulness of each encoding component in the dual encoding network, we conduct an ablation study as follows. Given varied combinations of the encoding components, seven models are trained. Table \ref{tab:ablation-encoding} summarizes the choices of video and text encodings and the corresponding performance.

Among the individual encoders, biGRU-CNN is found to be the most effective. 
As more encoding layers are included, the overall performance goes up. For the last four models which combine the output from previous layers, they all outperform the corresponding counterpart using the output of a specific layer. E.g., the model with Level 1 + 2 encoding strategy outperforms the ones with Level 1 or Level 2. The results suggest that features of different levels are complementary to each other. The full multi-level encoding setup, \ie Level 1 +2 + 3, performs the best.

We also investigate single-side encoding, that is, video-side multi-level encoding with mean pooling on the text side, and text-side multi-level encoding with mean pooling on the video side. These two strategies obtain SumR of 194.5 and 191, respectively. The lower scores justify the necessity of dual encoding. The result also suggests that video-side multi-level encoding is more beneficial.

\begin{figure*}[tb!]
\centering\includegraphics[width=2.0\columnwidth]{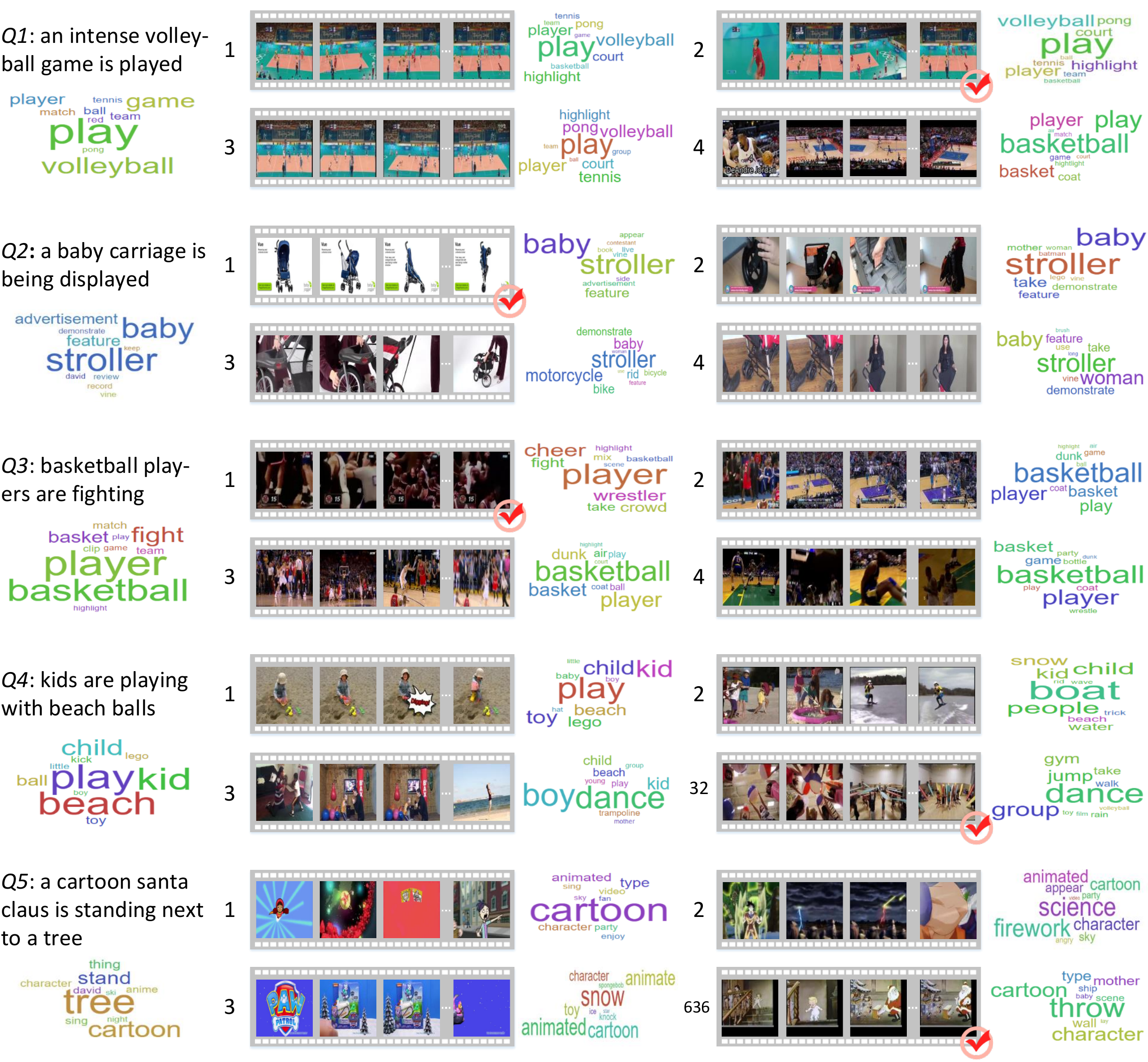}
\caption{\textbf{Selected examples of text-to-video retrieval by our model on MSR-VTT}. For each query, the top 3 ranked videos and the ground-truth video (marked with red ticks) are shown. In case the ground-truth video is among the top three, the fourth video will be included as well. By definition, each query has only one ground-truth video. Number on the left hand side of each video indicates the video's rank in the retrieval result. Below a specific query are its predicted concepts, visualized in the form of a tag cloud, bigger font meaning larger predicted scores. Next to the videos are their predicted concepts. Putting these tag clouds together helps us better understand the video retrieval results.}\label{fig:wordcloud}
\end{figure*}

\subsubsection{Hybrid Space versus Single Space}\label{sssec:hybrid-space}

In order to verify the effectiveness of the hybrid space, we have re-trained Dual Encoding with two alternative spaces, \ie fully latent space (the CVPR version~\cite{cvpr2019-dual-dong}) and fully concept space, respectively. Table \ref{tab:ablation-space} summarizes their performance on MSR-VTT.

Our model with the hybrid space consistently outperforms the other two counterparts with a clear margin, which shows the effectiveness of hybrid space for video-text retrieval.
Among them, although the model with the concept space is able to give some interpretation of the retrieval model, its performance is the worst. The latent space counterpart gives better performance than the concept space, but lacks interpretability. 
Moreover, simply increasing the dimensionality of the latent space, from 1,536 to 2,048, does not improve the performance. 
By contrast, the hybrid space strikes a proper balance between the retrieval performance and the interpretability.

To justify the necessity of the combined loss for concept space learning, we also report the performance of the hybrid space that excludes the triplet ranking loss from Eq. \ref{eq:p_loss}. This variant suffers a noticeable performance decrease in terms of SumR, from 211.7 to 183.3. The result shows the importance of considering the triplet ranking loss for learning a concept space that is beneficial for video-text matching.

\begin{table*} [tb!]
\renewcommand{\arraystretch}{1.2}
\caption{\textbf{Comparison with Transformer-based multi-modal methods for video retrieval by text}. Dataset: MSR-VTT. Even though Dual Encoding uses relatively simple 2D CNN video features, it outperforms MMT on Test1k-Miech \cite{miech2018learning}, while less effective on Test1k-Yu \cite{yu2018joint}. The inclusion of Transformer features as shown in Fig. \ref{fig:with_transformer} makes Dual Encoding top the performance table consistently on all the three test sets.}
\label{tab:dual_transformer}
\centering 
\scalebox{0.78}{
\begin{tabular}{@{}l*{18}{r} @{}}
\toprule
\multirow{2}{*}{\textbf{Method}}   && \multicolumn{5}{c}{\textbf{Official full-size test set \cite{xu2016msr}}} && \multicolumn{5}{c}{\textbf{Test1k-Miech \cite{miech2018learning}}} && \multicolumn{5}{c}{\textbf{Test1k-Yu~\cite{yu2018joint}}}  \\
 \cmidrule{3-7}  \cmidrule{9-13} \cmidrule{15-19} 
&& R@1 & R@5 & R@10 & Med r & mAP && R@1 & R@5 & R@10 & Med r & mAP && R@1 & R@5 & R@10 & Med r & mAP \\
\cmidrule{1-1}  \cmidrule{3-7}  \cmidrule{9-13} \cmidrule{15-19}
ActBERT \cite{zhu2020actbert}, HowTo100M pre-training   && - & - & - & - & - && - & - & - & - & - &&     8.6 & 23.4 & 33.1 & 36 & -\\
MMT \cite{gabeur2020multi}               && - & - & - & - & - && 20.3 & 49.1 & 63.9 & 6 & - &&     24.6 & 54.0 & 67.1 & 4 & -   \\
MMT \cite{gabeur2020multi}, HowTo100M pre-training  && - & - & - & - & - && - & - & - & - & - &&     26.6 & 57.1 & 69.6 & 4 & -  \\
\cmidrule{3-7}  \cmidrule{9-13} \cmidrule{15-19}
\textit{Dual Encoding}                   && 11.6 & 30.3 & 41.3 & 17 & 21.2 && 23.0 & 50.6 & 62.5 & \textbf{5} & 36.1 &&     21.1 & 48.7 & 60.2 & 6 & 33.6 \\
\textit{Dual Encoding}, with Transformer features only && 8.5 & 25.1 & 36.1 & 22 & 17.4 && 16.8 & 41.6 & 55.8 & 8 & 28.7 &&     15.0 & 40.7 & 54.2 & 8 & 27.6  \\
\textit{Dual Encoding}, with Transformer features included  && \textbf{13.3} & \textbf{34.0} & \textbf{45.7} & \textbf{13} & \textbf{23.8} && \textbf{25.7} & \textbf{52.9} & \textbf{66.4} & \textbf{5} & \textbf{38.9} && \textbf{32.4} & \textbf{62.3} & \textbf{72.8} & \textbf{3} & \textbf{46.0}  \\
\bottomrule
\end{tabular}
 }
\end{table*}

\textbf{Interpreting retrieval results with predicted concepts}. 
Fig. \ref{fig:wordcloud} shows some examples returned by our proposed dual encoding. 
Although only one correct video is annotated for each query, the top retrieved videos for Q1 (Query 1), Q2 and Q3 are typically relevant to the given query to some extent. 
In Q4, as the word \textit{beach} is used to describe the object \textit{ball}, the former is less important than the latter in this query. However, the predicted concept vector of Q4 shows that the model over emphasizes \textit{beach}, as visualized in the tag cloud. This explains that the top 2 retrieved videos are all about activities on beach. Meanwhile, for the truly relevant video, which is ranked at the position of 32, the predicted concepts are \textit{dance} and \textit{group}. Although these concepts are semantically relevant to the video content, they are irrelevant for the query.  
For Q5, concepts predicted our model, \eg \textit{cartoon} and \textit{tree}, are not precise enough to capture \textit{santa claus} the key role in the query. So our model also fails to answer this query. 
In general, we find concepts predicted by our dual encoding model  reasonable, and useful for understanding the retrieval model.

\begin{table} [tb!]
\renewcommand{\arraystretch}{1}
\caption{\textbf{Comparison with MMT in terms of model size and computation overhead at the inference stage}.  For each model, we measure the amount of FLOPs it takes to encode a given video-text pair. The computational cost of video feature extraction is excluded as that step is typically performed once in an offline mode. Numbers in parentheses indicate relative changes against MMT.}
\vspace{-3mm}
\label{tab:dual_transformer_complex}
\centering 
\scalebox{0.9}{
\begin{tabular}{@{}l*{2}{l} @{}}
\toprule
\multirow{2}{*}{\textbf{Model}}   & \multicolumn{2}{c}{\textbf{Model Complexity}} \\
 \cmidrule{2-3} 
 &  Parameters (M) & FLOPs (G) \\
\cmidrule{1-3} 
MMT               &  133.4 & 12.64 \\
\textit{Dual Encoding}            &   \textbf{~~95.9} ($\downarrow$ 28.1\%) & \textbf{~~3.64} ($\downarrow$ 71.2\%) \\
\bottomrule
\end{tabular}
 }
\end{table}

\subsubsection{Dual Encoding versus-and-with Transformers}\label{sssec:trans}
\textbf{Dual Encoding versus multi-modal Transformers}. We compare with two recent multi-modal Transformers, \ie ActBERT~\cite{zhu2020actbert} and MMT~\cite{gabeur2020multi}, that have been evaluated on specific test sets of MSR-VTT. 
As Table \ref{tab:dual_transformer} shows, Dual Encoding clearly outperforms ActBERT and is comparable to MMT in terms of the overall performance (better on Test1k-Miech and worse on Test1k-Yu). It is worth pointing out that MMT uses a diverse set of seven video features that describe varied aspects of the video content including motion, audio, scene, OCR, face, speech, and visual appearance features. By contrast, Dual Encoding uses only two regular 2D-CNN features. 
Moreover, on Test1k-Yu \cite{yu2018joint}, Dual Encoding was trained on the corresponding training set of 7,010 videos, whilst MMT was trained on an enlarged set of 9,000 videos.
We consider in this context that Dual Encoding compares favorably to MMT. 
In addition, Table~\ref{tab:dual_transformer_complex} shows a comparison concerning model complexity. Note that ActBERT is not included, as that method remains closed-source, making a precise estimation of its model complexity impossible. Dual Encoding is smaller (28.1\% less parameters) and much faster (71.2\% less FLOPs) than MMT.

\begin{figure}[tb!]
\centering\includegraphics[width=0.7\columnwidth]{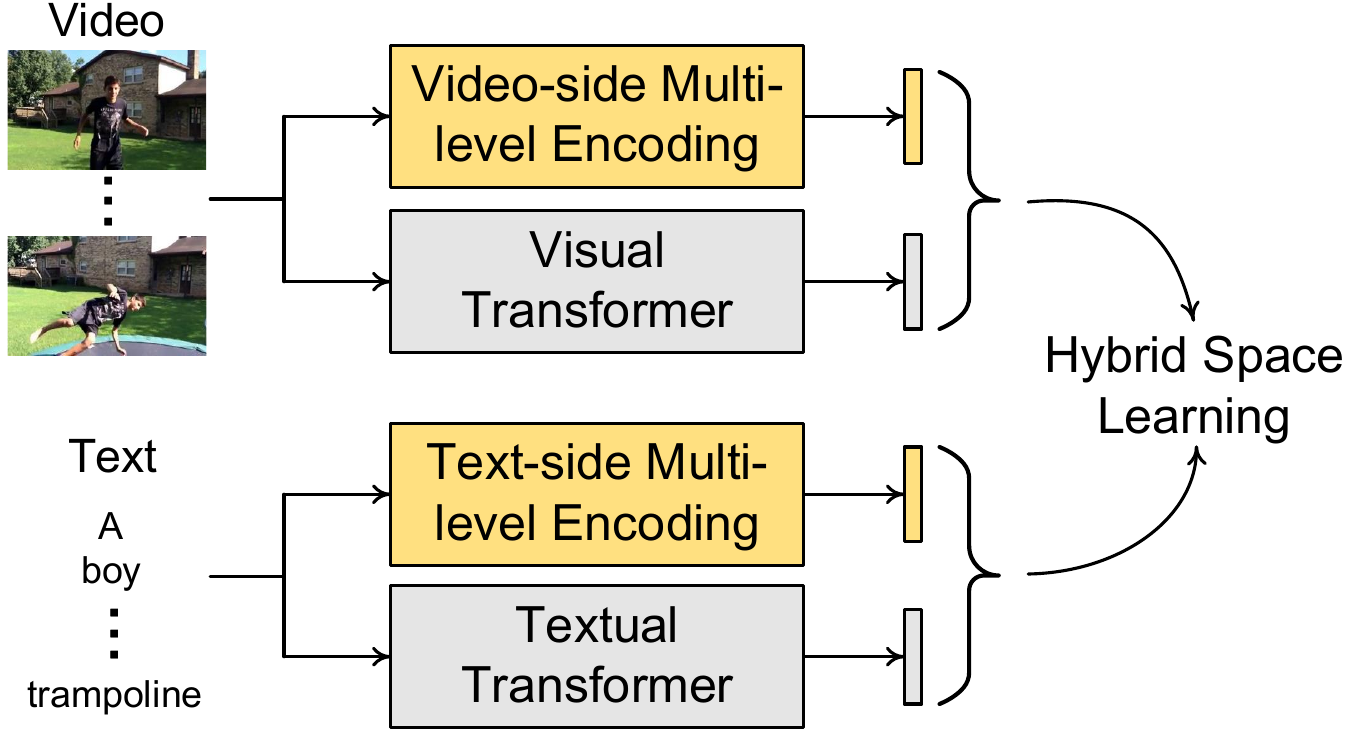}
\vspace{-3mm}
\caption{\textbf{Integrating pre-trained visual / textual Transformers into the Dual Encoding network}. The outputs of the two transformers are respectively concatenated with the video-side and text-side multi-level encodings in advance to hybrid space learning. }\label{fig:with_transformer}
\end{figure}

\textbf{Dual Encoding with Transformers}. Dual Encoding is a generic framework that allows pre-trained Transformers to be included with ease. As illustrated in Fig. \ref{fig:with_transformer},  we extend the Dual Encoding network by adopting MMT trained on the HowTo100M dataset \cite{miech2019howto100m} as the visual Transformer and BERT~\cite{bert-toolbox} trained on Wikipedia and book corpora~\cite{zhu2015aligning} as the textual Transformer. The two Transformers are integrated as is without fine tuning. Their outputs, \ie a 3,584-d video feature vector and a 1,024-d sentence vector, are concatenated with the video-side and text-side multi-level encodings, respectively, in advance to hybrid space learning.
As shown in Table \ref{tab:dual_transformer}, the inclusion of the Transformer features clearly and consistently boosts the retrieval performance on all the three test sets. Meanwhile, using the Transformer features alone is less effective than the previous Dual Encoding model. Hence, our multi-level encoding modules, while built upon the relatively simple mean feature pooling, GRU and 1D-CNN encoders, remain competitive and can be used together with the Transformers to maximize the retrieval performance.

\subsubsection{Multi-level Encoding for Image-Text Retrieval}

\begin{table} [tb!]
\renewcommand{\arraystretch}{1.2}
\caption{\textbf{Performance of image-text retrieval on Flickr30k and MSCOCO}. The proposed text-side multi-level encoding (MLE) is beneficial for VSE++~\cite{faghri2017vse}.}\vspace{-3mm}
\label{tab:image-text}
\centering 
\scalebox{0.9}{
\begin{tabular}{@{}l*{8}{r} @{}}
\toprule
\multirow{2}{*}{\textbf{Method}}   & \multicolumn{3}{c}{\textbf{Text-to-Image}} && \multicolumn{3}{c}{\textbf{Image-to-Text}} & \multirow{2}{*}{\textbf{SumR}}\\
 \cmidrule{2-4}  \cmidrule{6-8} 
& R@1 & R@5 & R@10 && R@1 & R@5 & R@10 & \\
\cmidrule{1-9}
\textit{On Flickr30k}                       &&& & &  &    &   \\
VSE++  & 23.1 & 49.2 & 60.7 &&     31.9 & 58.4 & 68.0 & 291.3 \\
VSE++, MLE & \textbf{24.7} & \textbf{52.3} & \textbf{65.1} &&      \textbf{35.1} & \textbf{62.2} & \textbf{71.3} & \textbf{310.7} \\
\cmidrule{1-9}
\textit{On MSCOCO}                         &&&  & &  &    &   \\
VSE++  & 33.7 & 68.8 & 81.0 && 43.6 & 74.8 & 84.6  &  389.6   \\
VSE++, MLE  &  \textbf{34.8} & \textbf{69.6} & \textbf{82.6} &&   \textbf{46.7} & \textbf{76.2} & \textbf{85.8} & \textbf{395.7}  \\
\bottomrule
\end{tabular}
 }
\end{table}

\textbf{Setup}.
We investigate if the VSE++ model \cite{faghri2017vse} can be improved in its original context of image-text retrieval, when replacing its textual encoding module, which is a GRU, by the proposed multi-level encoding module. To that end, we fix all other choices, adopting the exact evaluation protocol of \cite{faghri2017vse}. That is, we use the same data split, where the training / validation / test test has 30,000 / 1,000 / 1,000 images for Flickr30K, and 82,783 / 5,000 / 5,000 images for MSCOCO. We also use the same VGGNet feature provided by \cite{faghri2017vse}. Performance of $R@1$, $R@5$ and $R@10$ are reported. On MSCOCO, the results are reported by averaging over 5 folds of 1,000 test images.

\textbf{Results}.
Table \ref{tab:image-text} shows the performance of image-text retrieval on Flickr30k and MSCOCO. Integrating text-side multi-level encoding into VSE++ brings improvements on both datasets.  The results suggest that the proposed text-side multi-level encoding is also beneficial for VSE++ in its original context.

\section{Summary and Conclusions} \label{sec:conc}

For video retrieval by text, this paper proposes a Dual Encoding network with hybrid space learning. By jointly exploiting multiple encoding strategies at different levels, the proposed network encodes both videos and text into powerful dense representations. Followed by hybrid space learning, these representations can be transformed to perform sequence-to-sequence cross-modal matching effectively.
Extensive experiments on four video datasets, \ie MSR-VTT, TRECVID AVS 2016-2018, VATEX, and MPII-MD, support the following conclusions. Among the three levels of encoding, biGRU-CNN  that builds a 1D convolutional network on top of bidirectional GRU is the most effective when used alone. Video-side multi-level encoding is more beneficial when compared with its text-side counterpart. 
Our multi-level encoding modules also show competitive performance against recent multi-modal Transformers.
Compared with the widely used latent space learning, our hybrid space learning not only improves the retrieval performance but also enhances the interpretability of what the dual encoding network has learned.
For state-of-the-art performance, we recommend Dual Encoding with hybrid space learning for video-text matching.

\appendix

\begin{figure}[tb!]
\centering\includegraphics[width=0.55\columnwidth]{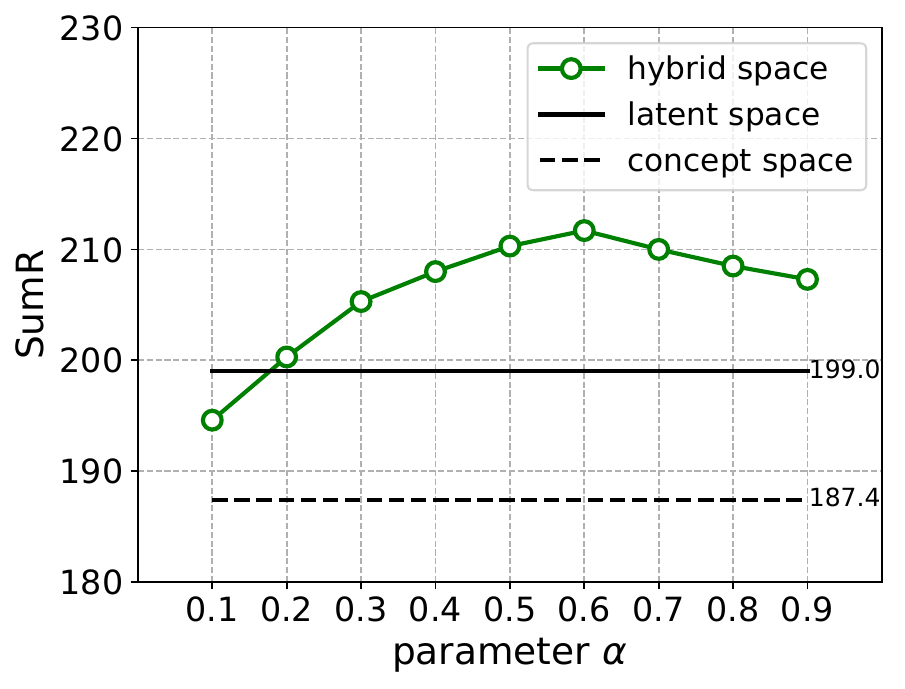}
\caption{\textbf{The influence of the parameter $\alpha$ of Eq. \ref{eq:final_simi} on our proposed model}.
Performance are evaluated on MSR-VTT.}\label{fig:hybrid_weight}
\end{figure}

\textbf{Effect of $\alpha$ on the retrieval performance}.
The influence of the hyper-parameter $\alpha$ in Eq. \ref{eq:final_simi} is studied as follows.
We try $\alpha$ with its value ranging from 0.1 to 0.9 with an interval of 0.1.
As shown in Fig. \ref{fig:hybrid_weight}, when the $\alpha$ is larger than 0.2, the performance of our model with the hybrid space are all over 200 on MSR-VTT, which consistently outperform the counterparts using the latent space or the concept space alone. The results show that our hybrid space is not very sensitive to this parameter.

\begin{figure}[tbp]
	\centering
	\subfigure[With $\mathcal{L}_{con,rank}$ \label{fig:noisy_dual_a}]{
		\includegraphics[width=0.48\columnwidth]{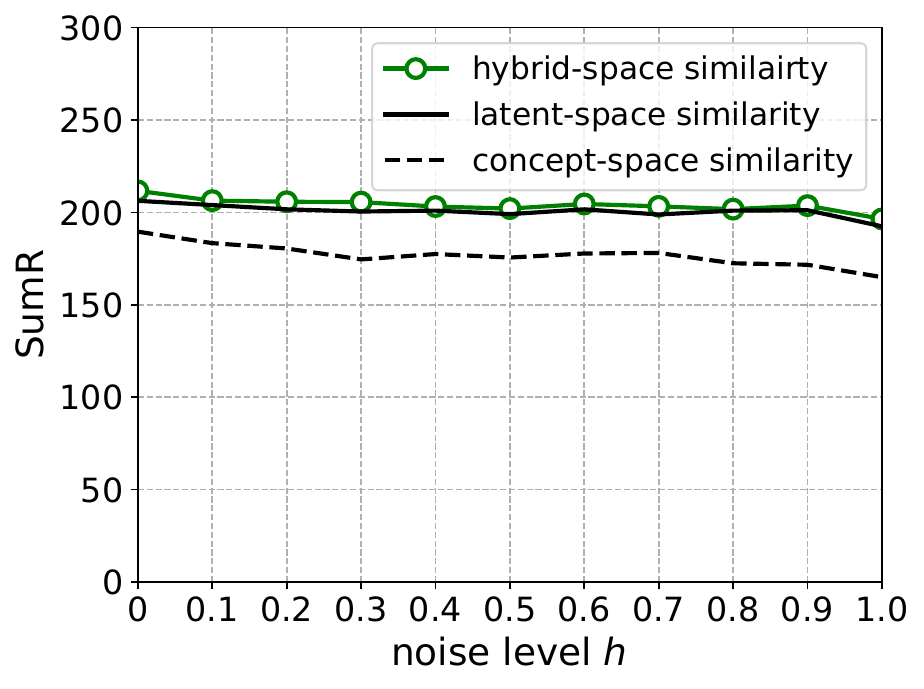}}
	\subfigure[Without $\mathcal{L}_{con,rank}$\label{fig:noisy_dual_b}]{
		\includegraphics[width=0.48\columnwidth]{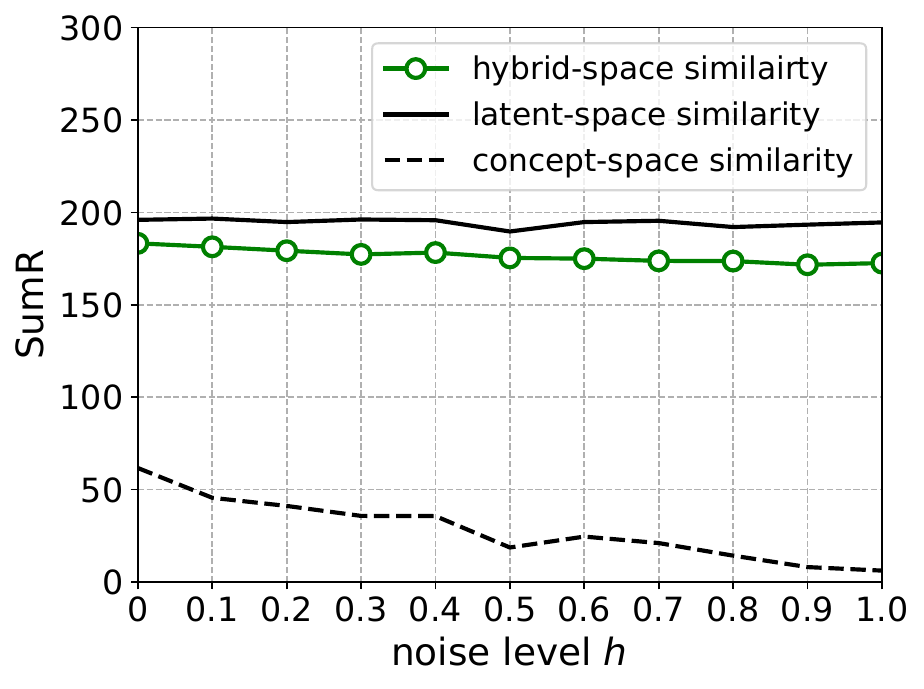}}
	\caption{\textbf{The influence of the concept annotation quality on the overall performance of text-to-video and video-to-text retrieval}. Per noise level, hybrid space learning is performed with  $\mathcal{L}_{con,rank}$ (a)  and without $\mathcal{L}_{con,rank}$ (b), respectively. The curve of ``hybrid-space similarity'' shows how the noise affects the model performance. To further reveal the effect of the noise on the individual spaces, we plot performance curves when using the latent-space similarity and concept-space similarity separately. The use of $\mathcal{L}_{con,rank}$ allows our model to be highly insensitive to the concept annotation quality.}
	\label{fig:noisy_dual}
\end{figure}

\textbf{Effect of the concept annotation quality on the retrieval performance}.
The concept annotations used to supervise the hybrid space learning process are automatically extracted from video descriptions. In order to explore how their quality affects the retrieval performance, we re-train the model using concept annotations with varied levels of simulated noise. Concretely, suppose a specific training video is originally associated with $q$ concepts. Given a noise level of $h \in \{0.1, 0.2,\cdots, 1\}$, we replace a random subset of $q \times h$ concepts in the original annotations by the same number of concepts selected from the remaining $K-q$ concepts (recall that K is the size of the concept vocabulary). A larger $h$ means lower annotation quality. 
As shown in Fig.~\ref{fig:noisy_dual_a}, the retrieval performance is hardly affected by the concept annotation quality at a large range of noise levels. We attribute this result to two factors. First, the hybrid-space similarity is dominated by the latent-space similarity, which is independent of concept annotations by definition. Hence, while the retrieval performance of the concept-space similarity degenerates as more noise is added, the performance of the latent-space similarity is largely stable. Second, the use of $\mathcal{L}_{con,rank}$   maintains to a large extent the effectiveness of the concept space for cross-modal matching. Note the extreme case of $h=1$ at the right end of the curves, where the concept annotations  become fully random and consequently the correspondence between the individual dimensions and concepts is completely lost. By minimizing $\mathcal{L}_{con,rank}$, the ``concept'' space remains a latent space that can be used for matching. Without the loss, the space will be meaningless, see Fig. \ref{fig:noisy_dual_b}.
Therefore, the retrieval performance of our model is highly insensitive to the concept annotation quality.
Although $\mathcal{L}_{bce}$ is relatively sensitive to the quality of annotation, we consider the loss necessary as it produces an auxiliary common space not only interpretable but also more complementary to the latent space, see the better performance of the hybrid space against the fully latent space in Table \ref{tab:ablation-space}.

\textbf{Efficiency Test}.
Recall that the dual encoding network is designed to represent both videos and sentences into a common space respectively. Once the network is trained, representing them in the common space can be performed independently. This means we can process large-scale videos offline and answer ad-hoc queries on the fly. 
Specifically, given a natural-sentence query, it takes approximately 0.2 seconds to retrieve videos from the largest IACC.3 dataset, which consists of 335,944 videos. The performance is tested on a normal computer with 64G RAM and a GTX 1080TI GPU. The retrieval speed is adequate for instant response.


%



\ifCLASSOPTIONcompsoc
  \section*{Acknowledgments}
\else
  \section*{Acknowledgment}
\fi

This work was supported by NSFC (No. 61902347, No. 61672523), BJNSF (No. 4202033), ZJNSF (No. LQ19F020002), the Public Welfare Technology Research Project of Zhejiang Province (No. LGF21F020010), the Fundamental Research Funds for the Central Universities and the Research Funds of Renmin University of China (No. 18XNLG19), and the  Alibaba-ZJU Joint Research Institute of Frontier Technologies.

\ifCLASSOPTIONcaptionsoff
  \newpage
\fi



\bibliographystyle{IEEEtran}
\bibliography{dual_encoding}
%



%




\end{document}